\documentclass[preprint, 12pt]{elsarticle}
\usepackage{lipsum}
\makeatletter

\makeatletter
\def\ps@pprintTitle{%
 \let\@oddhead\@empty
 \let\@evenhead\@empty
 \def\@oddfoot{}%
 \let\@evenfoot\@oddfoot}
\makeatother

\usepackage[utf8]{inputenc}
\usepackage{algpseudocode}
\usepackage{algorithm}
\usepackage[margin=1in]{geometry}
\usepackage{bold-extra}
\usepackage{adjustbox}
\usepackage{appendix}
\usepackage{lineno}

\usepackage{amssymb}
\usepackage{amsthm}
\usepackage{amsmath}
\usepackage{graphicx}
\usepackage{times}
\usepackage[titles]{tocloft}
\usepackage{enumerate}
\usepackage{bm}
\usepackage{sidecap}
\usepackage{epstopdf}
\usepackage{pdfpages}
\usepackage[colorinlistoftodos]{todonotes}
\usepackage{soul}
\usepackage{multicol}
\usepackage{float}
\usepackage{multibib}
\usepackage{geometry}
\usepackage[small,compact]{titlesec}
\usepackage[normal,bf,labelsep=colon,skip=2pt]{caption}
\usepackage{setspace}
\usepackage{subfig}
\usepackage{tocloft}
\usepackage{url}
\usepackage{floatflt}
\usepackage{array}
\usepackage{pgfgantt}
\usepackage{sidecap}
\usepackage{cite}
\usepackage[hidelinks]{hyperref}
\usepackage{multirow}

\numberwithin{equation}{section}
\numberwithin{figure}{section}
\numberwithin{table}{section}
\usepackage{tabularray}
\usepackage{diagbox}

\usepackage{xcolor}

\graphicspath{{./plots/}}

\newcommand*\diff{\mathop{}\!\mathrm{d}}

\begin{document}

\begin{frontmatter}

\title{CGKN: A Deep Learning Framework for Modeling Complex Dynamical Systems and Efficient Data Assimilation}

\author[1]{Chuanqi Chen}
\ead{cchen656@wisc.edu}
\author[2]{Nan Chen}
\ead{chennan@math.wisc.edu}
\author[2]{Yinling Zhang}
\ead{zhang2447@wisc.edu}
\author[1]{Jin-Long Wu\corref{cor1}} \ead{jinlong.wu@wisc.edu}
\cortext[cor1]{Corresponding author}

\address[1]{Department of Mechanical Engineering, University of Wisconsin–Madison, Madison, WI 53706}
\address[2]{Department of Mathematics, University of Wisconsin–Madison, Madison, WI 53706}

\begin{abstract}
Deep learning is widely used to predict complex dynamical systems in many scientific and engineering areas. However, the black-box nature of these deep learning models presents significant challenges for carrying out simultaneous data assimilation (DA), which is a crucial technique for state estimation, model identification, and reconstructing missing data. Integrating ensemble-based DA methods with nonlinear deep learning models is computationally expensive and may suffer from large sampling errors. To address these challenges, we introduce a deep learning framework designed to simultaneously provide accurate forecasts and efficient DA. It is named Conditional Gaussian Koopman Network (CGKN), which transforms general nonlinear systems into nonlinear neural differential equations with conditional Gaussian structures. CGKN aims to retain essential nonlinear components while applying systematic and minimal simplifications to facilitate the development of analytic formulae for nonlinear DA. This allows for seamless integration of DA performance into the deep learning training process, eliminating the need for empirical tuning as required in ensemble methods. CGKN compensates for structural simplifications by lifting the dimension of the system, which is motivated by Koopman theory. Nevertheless, CGKN exploits special nonlinear dynamics within the lifted space. This enables the model to capture extreme events and strong non-Gaussian features in joint and marginal distributions with appropriate uncertainty quantification. We demonstrate the effectiveness of CGKN for both prediction and DA on three strongly nonlinear and non-Gaussian turbulent systems: the projected stochastic Burgers–Sivashinsky equation, the Lorenz 96 system, and the El Ni\~no-Southern Oscillation. The results justify the robustness and computational efficiency of CGKN.
\end{abstract}
\begin{keyword}
Complex dynamical system \sep
Scientific machine learning \sep
Data assimilation \sep
Uncertainty quantification \sep
Koopman theory
\end{keyword}

\end{frontmatter}

\section{Introduction}
Complex dynamical systems are ubiquitous across a wide range of scientific and engineering fields, including climate science, geophysics, materials science, and neuroscience~\citep{jost2005dynamical, chen2023stochastic, wiggins2003introduction, majda2006nonlinear}. These systems are typically characterized by multiscale interactions, chaotic or turbulent dynamics, non-Gaussian distributions, intermittency, and extreme events~\citep{dijkstra2013nonlinear, palmer1993nonlinear, trenberth2015attribution, moffatt2021extreme, majda2003introduction, manneville1979intermittency, majda2018model}. For decades, modeling complex dynamical systems has primarily relied on governing equations derived from physical laws~\citep{arnold1974stochastic, van1976stochastic, kloeden1992stochastic, protter2005stochastic}, which require a deep understanding of the underlying physics. However, with the rapid growth of available data and computing resources, data-driven approaches have become increasingly promising for modeling these systems~\citep{karniadakis2021physics, duraisamy2019turbulence, brunton2022data}. These methods include sparse model identification~\citep{schaeffer2013sparse, brunton2016discovering, chen2023causality, chen2023ceboosting}, reduced-order models~\citep{majda2018strategies, carlberg2013gnat, noack2011reduced, xie2018data, maulik2020time, maulik2021reduced}, physics-informed machine learning~\citep{wang2017physics, karpatne2017theory, wu2018physics, raissi2019physics, sun2020surrogate, kashinath2021physics, zhou2021learning, han2023equivariant, yu2024learning}, stochastic parameterizations~\citep{mana2014toward, dawson2015simulating, schneider2021learning, wu2024learning}, and deep learning techniques~\citep{li2021fourier, lu2021learning, chen2024neural, dong2024data}. Among these, deep learning is particularly flexible in capturing nonlinear dynamics and can deliver superior forecast performance when sufficient training data is available, especially for systems with unknown governing equations. However, the highly nonlinear, black-box nature of deep learning models poses significant challenges for outer-loop scientific machine learning applications such as data assimilation.

Data assimilation (DA) \citep{majda2012filtering, law2015data, kalman1960new, kalman1961new, evensen1994sequential, bergemann2012ensemble} integrates observational data into models to enhance state estimation. DA was originally designed to improve the estimation of initial conditions for the real-time forecast. Nowadays, DA has wide applications in parameter estimation, model identification, optimal control, and reconstructing missing data. DA consists of a two-step procedure: a model forecast generates a statistical prediction, which is then refined in the analysis step by incorporating new observations, improving accuracy, and reducing uncertainty. Given the highly nonlinear nature of many complex dynamical systems, ensemble DA \citep{evensen2003ensemble, whitaker2002ensemble, burgers1998analysis, zhou2024bi} has become one of the most widely used methods in practice. Ensemble DA methods utilize an ensemble of realizations to approximate probabilistic distributions of target states and update the ensemble accordingly when new observations become available. When the governing equations of dynamical systems are unknown or too costly to simulate, machine learning is often employed to assist DA. First, machine learning can be used to build data-driven forecast models for DA \citep{gagne2020machine, wikner2021using, rasp2018deep, bonavita2020machine, farchi2021comparison, malartic2022state, farchi2023online, bocquet2020online}. Second, machine learning with observers enhances the prediction of complex systems, and the physics-informed machine learning can be used to assist in the design of observers~\citep{neofotistos2019machine, gottwald2021combining, vargas2024nonlinear}. Third, machine learning can streamline the entire DA process by developing end-to-end learning frameworks \citep{revach2022kalmannet, boudier2020dan, ouala2018neural, manucharyan2021deep, mou2023combining, boudier2023data}. A comprehensive review of machine learning with data assimilation can be found in~\citep{cheng2023machine}.

While separate end-to-end machine learning models can be developed for prediction and DA, it is practically useful to design a unified deep learning framework that simultaneously tackles both tasks. The framework aims to incorporate specific dynamical structures to explicitly capture some aspects of the underlying physics, distinguishing it from fully black-box models. The unique model structures can also be used to develop efficient forecasts and DA algorithms. 

In this paper, we develop a Conditional Gaussian Koopman Network (CGKN), which is a deep learning framework designed to address the challenges of modeling complex dynamical systems, performing efficient forecasts, and accelerating effective DA. The CGKN is structured as a neural stochastic differential equation. It operates in three main steps: (i) transforming the state of the original nonlinear system into that of a conditional Gaussian nonlinear system, (ii) learning the dynamics of the conditional Gaussian nonlinear system to enable rapid state forecast and DA, and (iii) transforming the forecast and DA results back into the original system. Notably, the transformations required in the first and third steps are often unknown \textit{a priori}, but they can be jointly learned along with the dynamics of the conditional Gaussian nonlinear system.

It is essential to highlight two crucial building blocks of the CGKN. First, the conditional Gaussian nonlinear system \citep{liptser2013statistics, chen2018conditional, chen2022conditional} encompasses a rich class of nonlinear and non-Gaussian stochastic differential equations (SDEs), embedding a conditional linear structure in the model. Many well-known complex nonlinear dynamical systems fit within the framework of conditional Gaussian nonlinear systems, which have been exploited to develop realistic systems and assist in creating numerous fast computational algorithms. Second, in contrast to standard Koopman theory, which transforms the original nonlinear system into a linear system \citep{koopman1931hamiltonian, budivsic2012applied, mezic2013analysis, pan2020physics, otto2021koopman, brunton2022modern, pan2023pykoopman, colbrook2024limits}, the CGKN aims to transform the original system into a specific nonlinear system characterized by conditional linear structures. In this context, the dynamics of the unobserved states become linear when the observed state is known, while the interactions among different variables remain highly nonlinear. This intrinsic nonlinear dynamics enables the model to capture extreme events and strong non-Gaussian features in both joint and marginal distributions, with appropriate uncertainty quantification. Notably, the conditional linear structure allows the development of analytic formulae for DA to estimate unobserved states given the time series data of the observed states. With these closed analytic formulae, the quantification of the DA performance can be easily integrated into the loss function. It serves as an additional training criterion for the deep learning framework that is both efficient and robust. The analytic formulae significantly enhance the speed of DA and eliminate the need for empirical tunings as in the ensemble methods. Additionally, it reduces the randomness and inaccuracies stemming from inevitable large sampling errors. This approach has recently been explored in \citep{chen2024cgnsde} and holds the potential to enhance not only DA but also the quality of the learned model, particularly regarding the probabilistic relationship between observed and unobserved states.
 
The rest of this paper is organized as follows. Section~\ref{Sec:Methodology} presents the CGKN framework. The numerical experiments are presented in Section~\ref{Sec:NumericalExperiments} includes three numerical examples of applying the CGKN to complex turbulent systems with intermittency and extreme events that demonstrate the effectiveness and efficiency of CGKN. The paper is concluded in Section~\ref{Sec:Conclusion}.

\section{Methodology}
\label{Sec:Methodology}

Considering a complex dynamical system in the general form:
\begin{equation}
\begin{aligned}
\label{eq:DS}
    \frac{\mathrm{d} \mathbf{u}}{\mathrm{d} t} = \mathcal{M}(\mathbf{u}),
\end{aligned}
\end{equation}
where $\mathbf{u} \in \mathbb{R}^{d_{\mathbf{u}}}$ denotes the system state and $\mathcal{M}: \mathbb{R}^{d_{\mathbf{u}}} \mapsto \mathbb{R}^{d_{\mathbf{u}}}$ is a real vector-valued analytic function that describes the system dynamics. The system state $\mathbf{u}$ can be a high-dimensional vector and the dynamics function $\mathcal{M}$ can be highly nonlinear. The system in the form of Eq.~\eqref{eq:DS} can encompass a wide range of complex dynamical behaviors including multi-scale dynamics, chaos, turbulence, stochasticity, non-Gaussian statistics, intermittency, and extreme events. In this work, we focus on partially observed dynamical systems, with observed state variables denoted as $\mathbf{u}_1 \in \mathbb{R}^{d_{\mathbf{u}_1}}$ and the unobserved state variables denoted as $\mathbf{u}_2 \in \mathbb{R}^{d_{\mathbf{u}_2}}$. The unobserved states are the states that are not directly measured and to be inferred from data of observed states through techniques of data assimilation (DA). To explicitly discriminate the observed state variables $\mathbf{u}_1$ and unobserved state variables $\mathbf{u}_2$, the true system in Eq.~\eqref{eq:DS} is rewritten as:
\begin{equation}
    \begin{aligned}
        \label{eq:DS_u1u2}
        &\frac{\mathrm{d} \mathbf{u}_1}{\mathrm{d} t} = \mathcal{M}_1(\mathbf{u}_1, \mathbf{u}_2) ,\\
        &\frac{\mathrm{d} \mathbf{u}_2}{\mathrm{d} t} = \mathcal{M}_2(\mathbf{u}_1, \mathbf{u}_2),
    \end{aligned}
\end{equation}
where $[\mathbf{u}_1; \mathbf{u}_2]=\mathbf{u}$, $\mathcal{M}_1:\mathbb{R}^{d_{\mathbf{u}}} \mapsto \mathbb{R}^{d_{\mathbf{u}_1}}$ and $\mathcal{M}_2: \mathbb{R}^{d_{\mathbf{u}}} \mapsto \mathbb{R}^{d_{\mathbf{u}_2}}$ are real vector-valued analytic functions. The notation $[\cdot; \cdot]$ is used to denote the vertical concatenation of two column vectors. The DA task targets at the conditional distribution $p(\mathbf{u}_2(t)|\mathbf{u}_1(s), s\leq t)$, where the trajectory of $\mathbf{u}_1$ up to time $t$, namely $\mathbf{u}_1(s\leq t)$, is the observations while the state of $\mathbf{u}_2$ at $t$ needs to be estimated \citep{majda2012filtering, kalman1961new, law2015data}.

As a fundamental component of deep learning, a neural network is a computational model inspired by the structure of biological neural networks, consisting of hierarchical layers of neurons, each computing a weighted sum of inputs followed by a non-linear activation function. The universal approximation theorem~\citep{cybenko1989approximation, hornik1989multilayer, rumelhart1986learning} states that a neural network with at least one hidden layer, a sufficient number of neurons, and a non-linear activation function can approximate any continuous function. Therefore, both $\mathcal{M}_1$ and $\mathcal{M}_2$, which can be highly nonlinear functions, can be approximated by neural networks according to universal approximation theory. With a sufficient amount of data, the approximation of neural networks can be quite accurate, leading to a good performance of state forecasts. However, the DA for the black-box neural-networks-based model would require techniques such as the ensemble-based method, which are computationally expensive, could suffer from sampling errors, and often require empirical tuning.

\begin{figure}[H]
    \centering
    \includegraphics[width=\linewidth]{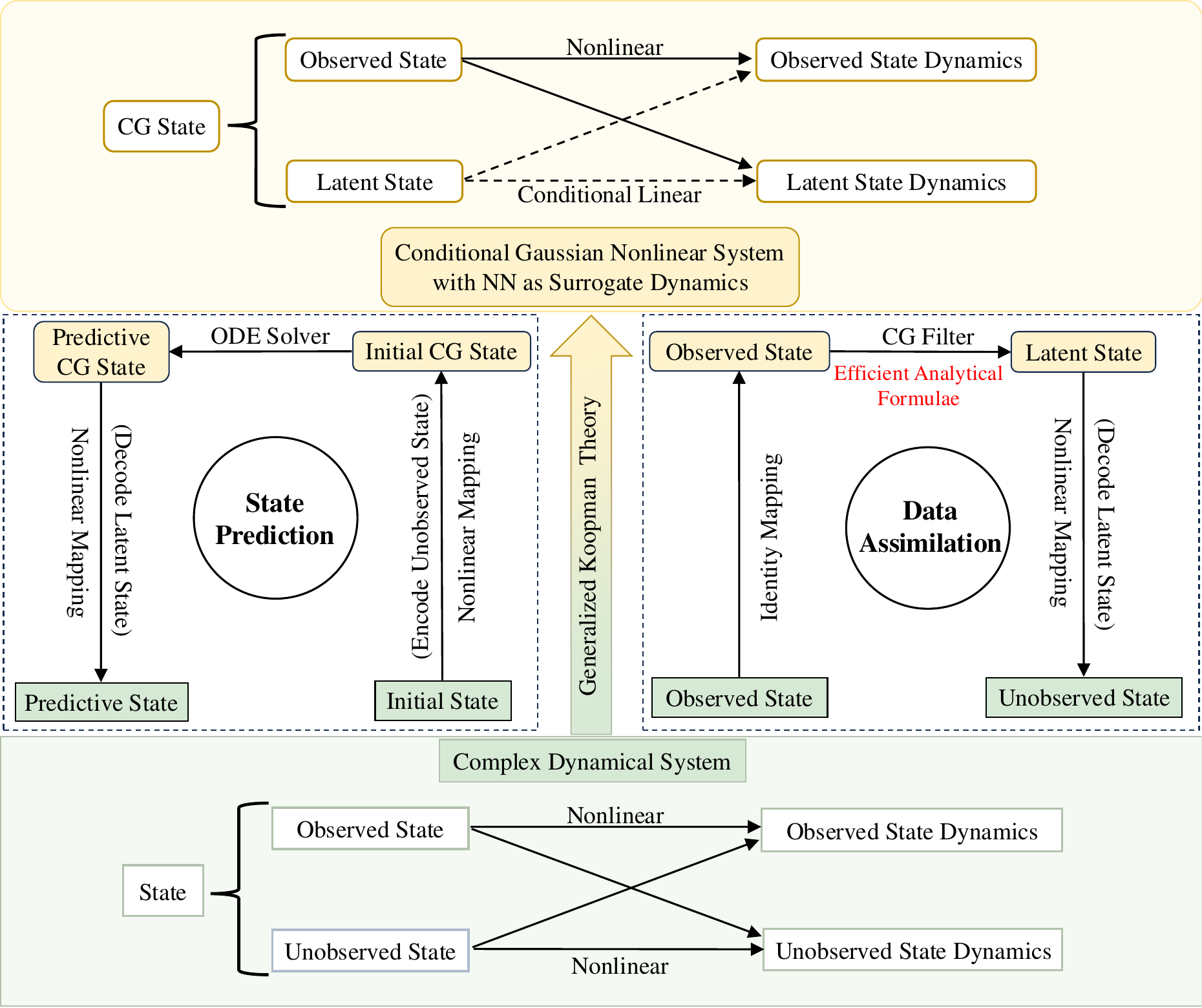}
    \caption{Schematic diagram of conditional Gaussian Koopman network (CGKN) for a partially observed system. The complex nonlinear dynamical system is transformed into the conditional Gaussian nonlinear system via a generalized usage of Koopman theory, which maps between the unobserved state variables and the latent state variables featured by conditional linear dynamics. The unknown dynamics of the conditional Gaussian nonlinear system are constructed by neural networks and jointly learned with the unknown nonlinear mappings. The analytic DA formulae for the conditional Gaussian nonlinear system can significantly accelerate DA process and also allow a computationally affordable DA loss to be incorporated into the CGKN training process. The trained CGKN model can be used for state forecasts and efficient DA, for which the computations are mainly performed in the conditional Gaussian state space and then mapped back to the original state space.}
    \label{fig:SchematicDiagram1}
\end{figure}

In this work, we develop the Conditional Gaussian Koopman Network (CGKN) for both state forecast and efficient DA. Instead of directly building a data-driven model for state variables $\mathbf{u}_1$ and $\mathbf{u}_2$, the proposed framework leverages a generalized version of the Koopman theory and exploits a proper transformation to map the unobserved state variables $\mathbf{u}_2$ to latent state variables $\mathbf{v}$, such that those latent variables are conditionally linear in the system dynamics. Therefore, the observed state variables $\mathbf{u}_1$ and the latent state variables $\mathbf{v}$ can be modeled by a conditional Gaussian nonlinear system~\citep{chen2018conditional, chen2022conditional, liptser2013statistics}, which encompasses many nonlinear models across various disciplines with many applications in natural science and engineering \citep{majda2009mathematical, grooms2014stochastic, branicki2013dynamic, keating2012new, chen2014information, chen2015noisy, chen2017beating}. Importantly, the conditional Gaussian nonlinear system facilitates the development of closed analytic formulae for DA. The analytic formulae provide more efficient DA than those techniques developed for nonlinear models and also enable the incorporation of DA performance into the loss function when training the CGKN model. 
It is essential to highlight that the governing equations of complex dynamical systems do not need to be known as CGKN is a data-driven method based on deep learning. If the governing equations are available, CGKN can be learned from the synthetic data generated from the simulation of the complex system and then provides efficient data assimilation. Alternatively, even if the governing equations are unknown, CGKN can still be learned using real data, for which this work assumes the data of unobserved states is accessible at the offline training stage. With a pre-trained CGKN, the inference processes, including both state prediction and efficient DA, do not require data from unobserved states.
A schematic overview of the CGKN framework, including its application in state forecast and DA, is presented in Fig.~\ref{fig:SchematicDiagram1}.

\subsection{Koopman Theory}

Koopman theory~\citep{koopman1931hamiltonian} provides a linear perspective on nonlinear dynamical systems in the form of Eq.~\eqref{eq:DS} by describing the system in terms of the evolution of observable of the system state. The observable $h: \mathbb{R}^{d{_\mathbf{u}}} \mapsto \mathbb{R}$ is a function in the Hilbert space operating on the system state $\mathbf{u}$. For the complex dynamical system in Eq.~\eqref{eq:DS}, the state transition operator $\mathcal{G}^{\Delta t}: \mathbb{R}^{d{_\mathbf{u}}} \mapsto \mathbb{R}^{{d_\mathbf{u}}}$ over time interval $\Delta t$ is defined as:
\begin{equation}
    \mathcal{G}^{\Delta t}\big(\mathbf{u}(t)\big) := \mathbf{u}(t) + \int_{t}^{t+\Delta t} \mathcal{M}\big(\mathbf{u}(\tau)\big)d \tau.
\end{equation}

The Koopman operator $\mathcal{K}^{\Delta t}$ is a linear operator acting on observable function $h$ which is defined as
\begin{equation}
    \mathcal{K}^{\Delta t}(h) := h \circ \mathcal{G}^{\Delta t},
\end{equation} where $\circ$ denotes function composition.

Furthermore, the Koopman infinitesimal generator $\mathcal{L}$ is defined as:
\begin{equation}
    \mathcal{L} := \lim_{\Delta t \rightarrow 0} \frac{\mathcal{K}^{\Delta t}(h) - h}{\Delta t},
\end{equation} which characterizes the $\mathcal{L}$ as the rate of change of the observable $h$ since $\mathcal{K}^{\Delta t}(h(\mathbf{u}(t))) = h \circ \mathcal{G}^{\Delta t}(\mathbf{u}(t)) = h(\mathbf{u}(t+\Delta t))$.

Therefore, the evolution of observable $h$ is governed by the Koopman infinitesimal generator $\mathcal{L}$ which is a linear operator:
\begin{equation}
    \frac{\partial h}{\partial t} = \mathcal{L} h.
\end{equation}

In real application, the infinite-dimensional linear dynamical system and the linear operator $\mathcal{L}$ are approximated by the finite-dimensional representations:
\begin{equation}
\label{eq:LDS}
    \frac{\mathrm{d} \mathbf{v}}{\mathrm{d} t} = \mathbf{A}\mathbf{v},
\end{equation} where $\mathbf{v} \in \mathbb{R}^{d_{\mathbf{v}}}$ is a finite-dimensional vector that approximates the infinite-dimensional function $h$ and the matrix $\mathbf{A} \in \mathbb{R}^{d_{\mathbf{v}} \times d_{\mathbf{v}}}$ is the linear dynamics of this discrete linear dynamical system which approximates the linear operator $\mathcal{L}$.

\subsection{Conditional Gaussian Koopman Network (CGKN)}
\label{SSec:CGKN}

In this work, a generalized application of Koopman theory has been employed to linearize the unobserved states $\mathbf{u}_2$ while retaining the nonlinearity of observed states $\mathbf{u}_1$ for the partially observed dynamical systems in Eq.~\eqref{eq:DS_u1u2}. The resulting conditional linear dynamical system is called the Conditional Gaussian Nonlinear System (CGNS), which is written as:
\begin{equation}
\begin{aligned}
\label{eq:CGKN}
\frac{\mathrm{d} \mathbf{u}_1}{\mathrm{d} t} &= \mathbf{f}_1(\mathbf{u}_1) + \mathbf{g}_1(\mathbf{u}_1)\mathbf{v} + \boldsymbol{\sigma}_1\dot{\mathbf{W}}_1,\\
\frac{\mathrm{d} \mathbf{v}}{\mathrm{d} t} &= \mathbf{f}_2(\mathbf{u}_1) + \mathbf{g}_2(\mathbf{u}_1)\mathbf{v} + \boldsymbol{\sigma}_2\dot{\mathbf{W}}_2,
\end{aligned}
\end{equation}
where $\mathbf{f}_1: \mathbb{R}^{d_{\mathbf{u}_1}} \mapsto \mathbb{R}^{d_{\mathbf{u}_1}}$, $\mathbf{g}_1: \mathbb{R}^{d_{\mathbf{u}_1}} \mapsto \mathbb{R}^{d_{\mathbf{u}_1}\times d_{\mathbf{v}}}$, $\mathbf{f}_2: \mathbb{R}^{d_{\mathbf{u}_1}} \mapsto \mathbb{R}^{d_{\mathbf{v}}}$, and $\mathbf{g}_2: \mathbb{R}^{d_{\mathbf{u}_1}} \mapsto \mathbb{R}^{d_{\mathbf{v}}\times d_{\mathbf{v}}}$ are nonlinear functions of $\mathbf{u}_1$ to be learned from data. $\mathbf{W}_1 \in \mathbb{R}^{d_{\mathbf{u}_1}}$ and $\mathbf{W}_2 \in \mathbb{R}^{d_{\mathbf{v}}}$ are independent Wiener processes with $\boldsymbol{\sigma}_1 \in \mathbb{R}^{d_{\mathbf{u}_1} \times d_{\mathbf{u}_1}}$ and $\boldsymbol{\sigma}_2 \in \mathbb{R}^{d_{\mathbf{v}} \times d_{\mathbf{v}}}$ as the noise coefficients. $\mathbf{v} = \boldsymbol{\varphi}(\mathbf{u}_2) \in \mathbb{R}^{d_\mathbf{v}}$ is called latent states with a nonlinear encoder $\boldsymbol{\varphi}:\ \mathbb{R}^{d_{\mathbf{u}_2}} \mapsto \mathbb{R}^{d_\mathbf{v}}$. It is worth noting that the conditional distribution of the latent variables $p(\mathbf{v}(t)|\mathbf{u}_1(s),s \leq t)$ is Gaussian, due to the conditional linear structure in Eq.~\eqref{eq:CGKN}, which facilitates efficient analytic formulae of DA with more details discussed in Section~\ref{SSec:CGKN_DA}.  With the state prediction and DA results of $[\mathbf{u}_1; \mathbf{v}]$, the corresponding original unobserved state variables can be obtained via $\mathbf{u}_2=\boldsymbol{\psi}(\mathbf{v})$ with the nonlinear decoder $\boldsymbol{\psi}:\ \mathbb{R}^{d_\mathbf{v}} \mapsto \mathbb{R}^{d_{\mathbf{u}_2}}$. The temporal derivative of the system states $\mathrm{d} \mathbf{u}/\mathrm{d} t$ is sometimes denoted as $\dot{\mathbf{u}}$ for short. The encoder $\boldsymbol{\varphi}$, decoder $\boldsymbol{\psi}$, and the four nonlinear functions $\mathbf{f}_1$, $\mathbf{g}_1$, $\mathbf{f}_2$, and $\mathbf{g}_2$ are parameterized by neural networks and are jointly learned from data $\{\mathbf{u}^{\star}(t_n)\}_{n=0}^N$. The neural networks that approximate the four nonlinear functions are collectively referred to as $\boldsymbol{\eta}$. 
 
\begin{figure}[H]
    \centering
    \includegraphics[width=\textwidth]{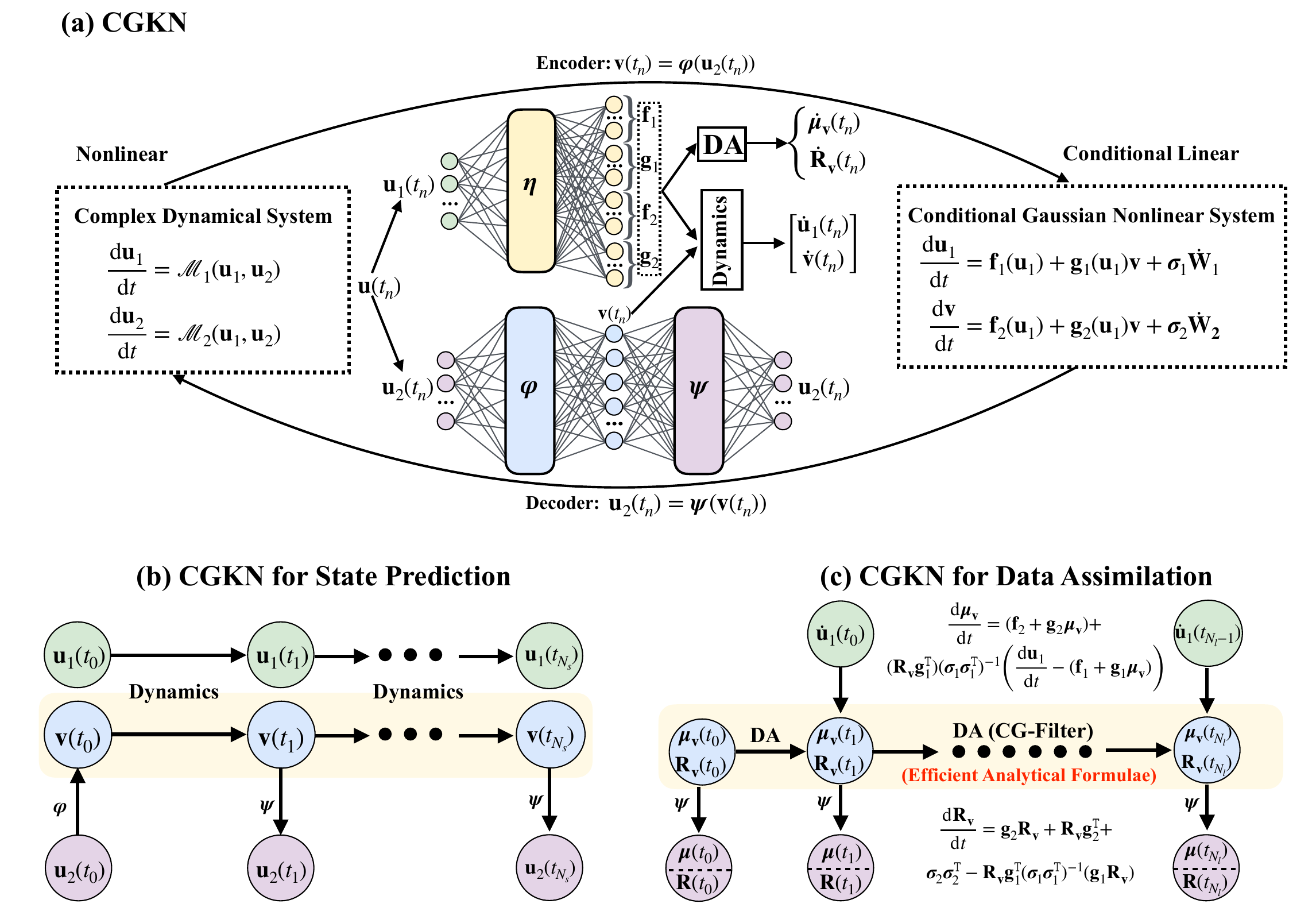}
    \caption{The architecture of CGKN and its applications for state prediction and data assimilation. The CGKN approximates the true nonlinear complex dynamical system by a modeled system with a conditional linear structure, which is also known as a conditional Gaussian nonlinear system. The conditional linear structure is enabled by a proper choice of latent state variables $\mathbf{v}$ as illustrated in (a). The nonlinear mappings between the original state variables $\mathbf{u}_2$ and the latent state variables $\mathbf{v}$ are achieved via the encoder $\boldsymbol{\varphi}$ and decoder $\boldsymbol{\psi}$, which are jointly learned with the unknown functions $\mathbf{f}_1$, $\mathbf{g}_1$, $\mathbf{f}_2$, $\mathbf{g}_2$ of the conditional Gaussian nonlinear system. The pre-trained CGKN can be used for state prediction and efficient data assimilation as illustrated in (b) and (c).}
    \label{fig:SchematicDiagram2}
\end{figure}

The CGKN comprises an encoder $\boldsymbol{\varphi}$, a decoder $\boldsymbol{\psi}$, and sub-networks $\boldsymbol{\eta}$. More specifically, the architecture of CGKN consists of (i) an encoder that maps the original unobserved system state variables to the latent state variables of a conditional Gaussian nonlinear system, (ii) neural networks that approximate the unknown terms in the dynamics of the conditional Gaussian nonlinear system, and (iii) a decoder that transforms the latent state variables of the conditional Gaussian nonlinear system back to the original unobserved system state variables. The architecture of CGKN is illustrated in Fig.~\ref{fig:SchematicDiagram2}(a). As a high-level deep learning framework, CGKN can integrate various neural network architectures, including fully-connected neural networks, residual neural networks, convolutional neural networks, and recurrent neural networks for encoder $\boldsymbol{\varphi}$, decoder $\boldsymbol{\psi}$, and sub-nets $\boldsymbol{\eta}$. In this work, the fully-connected neural networks are used to demonstrate the effectiveness and capability of CGKN.

It is worth noting that the CGKN framework is also applicable to another general type of a partially observed system, which consists of a dynamical model for the whole system state and an observation mapping that partially extracts observable information from the system state. This type of partially observed system can be formalized in the same form as shown in Eq.~\eqref{eq:DS_u1u2} and then modeled by the CGKN in Eq.~\eqref{eq:CGKN}. This is achieved by retaining the dynamical model and modeling the dynamics of the observations in relation to the system states, where the system state corresponds to the unobserved states $\mathbf{u}_2$ and the observations correspond to the observed states $\mathbf{u}_1$ in Eq.~\eqref{eq:CGKN}. If the dynamics of observations do not exist or if the observation time interval is large, a discrete-time version of CGKN can be employed.

\subsubsection{CGKN for State Prediction}
\label{SSec:CGKN_StatePrediction}

The CGKN model can be used for state prediction given an initial state $\mathbf{u}^{\star}(t_0) = [\mathbf{u}_1^{\star}(t_0); \mathbf{u}^{\star}_2(t_0)]$. Firstly, the unobserved state variables $\mathbf{u}^{\star}_2(t_0)$ are transformed into latent state variables $\mathbf{v}^{\star}(t_0)$ via an encoder $\boldsymbol{\varphi}$, which leads to the initial state of conditional Gaussian nonlinear system $\mathbf{u}_{\textrm{CG}} = [\mathbf{u}_1; \mathbf{v}]$ for Eq.~\eqref{eq:CGKN}:

\begin{equation}
\label{eq:u_CG_t0}
\begin{aligned}
\mathbf{u}_{\textrm{CG}}^{\star}(t_0)  =
    \begin{bmatrix}
        \mathbf{u}^{\star}_1(t_0) \\
        \mathbf{v}^{\star}(t_0)
    \end{bmatrix} =
        \begin{bmatrix}
        \mathbf{u}^{\star}_1(t_0) \\
        \boldsymbol{\varphi}(\mathbf{u}^{\star}_2(t_0))
    \end{bmatrix}.
\end{aligned}
\end{equation}

Then, the state prediction of the conditional Gaussian nonlinear system at a future time $t_n$ can be obtained by:
\begin{equation}
\label{eq:u_CG_Pred}
\begin{aligned}
\mathbf{u}_{\textrm{CG}}(t_n)  =
    \begin{bmatrix}
        \mathbf{u}_1(t_n) \\
        \mathbf{v}(t_n)
    \end{bmatrix} = \begin{bmatrix}
        \mathbf{u}^{\star}_1(t_0) + \int_{t_0}^{t_n} (\mathbf{f}_1(\mathbf{u}_1(t)) + \mathbf{g}_1(\mathbf{u}_1(t))\mathbf{v}) \mathrm{d} t \\
        \mathbf{v}^{\star}(t_0) + \int_{t_0}^{t_n} (\mathbf{f}_2(\mathbf{u}_1(t)) + \mathbf{g}_2(\mathbf{u}_1(t))\mathbf{v}) \mathrm{d} t
    \end{bmatrix}.
\end{aligned}
\end{equation}

Lastly, the predictive latent state variables $\mathbf{v}(t_n)$ are transformed back into the original unobserved state variables $\mathbf{u}_2(t_n)$ via a decoder $\boldsymbol{\psi}$:
\begin{equation}
\label{eq:u_Pred}
\begin{aligned}
\mathbf{u}(t_n)  =
    \begin{bmatrix}
        \mathbf{u}_1(t_n) \\
        \mathbf{u}_2(t_n)
    \end{bmatrix} =
        \begin{bmatrix}
        \mathbf{u}_1(t_n) \\
        \boldsymbol{\psi}(\mathbf{v}(t_n))
    \end{bmatrix}.
\end{aligned}
\end{equation}

In this work, the encoder $\boldsymbol{\varphi}$, the decoder $\boldsymbol{\psi}$, and the functions $\mathbf{f}_1$, $\mathbf{g}_1$, $\mathbf{f}_2$, $\mathbf{g}_2$ are jointly learned to ensure a good state prediction performance, i.e., the difference between the state prediction $\mathbf{u}(t_n)$ and the true system state $\mathbf{u}^{\star}(t_n)$ is small for a series of $t_n$. This is achieved by incorporating the state prediction performance into the loss function, and more details of the loss function for training a CGKN model are summarized in Section \ref{SSec:CGKN_Learning}. An illustration of CGKN for state prediction is shown in Fig.~\ref{fig:SchematicDiagram2}(b).

If the initial value of unobserved states $\mathbf{u}^{\star}_2(t_0)$ are not given, the estimated unobserved states via data assimilation after a warm-up period can substitute for the unknown true initial value. The CGKN for DA is introduced in Section~\ref{SSec:CGKN_DA}. Forecasting the entire system state using data from partially observed states is one of the features of CGKN, and it is worth noting that the performance of the state forecast will thus depend on the quality of DA estimated unobserved states.

\subsubsection{CGKN for Data Assimilation}
\label{SSec:CGKN_DA}

In addition to the state prediction, the CGKN model can also be used for efficient DA. This is achieved by the conditional linear structure in Eq.~\eqref{eq:CGKN} and accounts for one of the key highlights in the proposed CGKN framework. More specifically, despite the highly nonlinear dynamics of the coupled system, the conditional distribution $p(\mathbf{v}(t)|\mathbf{u}_1(s), s\leq t) \sim \mathcal{N}(\boldsymbol{\mu}_{\mathbf{v}}(t)), \mathbf{R}_{\mathbf{v}}(t))$ is Gaussian. The mean $\boldsymbol{\mu}_{\mathbf{v}}$ and covariance $\mathbf{R}_{\mathbf{v}}$ of this conditional Gaussian distribution can be solved by closed analytic formulae:

\begin{equation}
\begin{aligned}
\label{eq:CGKN_CGF}
&\frac{\mathrm{d} \boldsymbol{\mu}_{\mathbf{v}}}{\mathrm{d} t} = (\mathbf{f}_2 + \mathbf{g}_2\boldsymbol{\mu}_{\mathbf{v}}) + (\mathbf{R}_{\mathbf{v}}\mathbf{g}_1^\mathtt{T})(\boldsymbol{\sigma}_1\boldsymbol{\sigma}_1^\mathtt{T})^{-1} \left(\frac{\mathrm{d} \mathbf{u}_1}{\mathrm{d} t}-(\mathbf{f}_1 + \mathbf{g}_1\boldsymbol{\mu}_{\mathbf{v}})\right),\\
&\frac{\mathrm{d} \mathbf{R}_\mathbf{v} }{\mathrm{d} t} = \mathbf{g}_2\mathbf{R}_\mathbf{v} + \mathbf{R}_\mathbf{v}\mathbf{g}_2^\mathtt{T} + \boldsymbol{\sigma}_2\boldsymbol{\sigma}_2^\mathtt{T} - \mathbf{R}_\mathbf{v}\mathbf{g}_1^\mathtt{T}(\boldsymbol{\sigma}_1\boldsymbol{\sigma}_1^\mathtt{T})^{-1}(\mathbf{g}_1\mathbf{R}_\mathbf{v}).
\end{aligned}
\end{equation}
Given the data of observed state variables $\{\mathbf{u}_1^{\star}(t_n)\}_{n=0}^N$, the posterior mean of latent state variables $\{\boldsymbol{\mu}_{\mathbf{v}}(t_n)\}_{n=0}^N$ in Eq.~\eqref{eq:CGKN} can be estimated based on the analytic formulae in Eq.~\eqref{eq:CGKN_CGF}. Similar to the latent state variables, the posterior mean $\boldsymbol{\mu}_{\mathbf{v}}$ of the latent state variables can also be transformed back to the mean estimation $\boldsymbol{\mu}$ of the original unobserved state variables $\mathbf{u}_2$ via the decoder $\boldsymbol{\psi}$, i.e., $\boldsymbol{\mu}(t_n) \approx \boldsymbol{\psi}(\boldsymbol{\mu}_{\mathbf{v}}(t_n))$. Although this direct nonlinear transformation of the mean values is efficient, it is generally not valid for an arbitrary nonlinear decoder. However, its validity can be greatly enhanced by a trainable decoder with the DA loss being considered. In the DA process, the exact initial condition or initial distribution of the DA solution is usually unknown. As a result, it takes some time for the DA solution to stabilize and adjust to eliminate the inconsistency from the initialization. During this initial period of DA, which is known as the warm-up period, the estimated states may deviate from the actual values. Consequently, warm-up period is crucial for ensuring that DA produces stable and accurate estimates for unobserved states before proceeding with any analysis or decision-making. Therefore, the steps of DA for the warm-up period are omitted in the evaluation of DA loss for training the CGKN. More details about the DA loss are discussed in Section~\ref{SSec:CGKN_Learning}. An illustration of CGKN for DA is shown in Fig.~\ref{fig:SchematicDiagram2}(c).

The effectiveness of DA depends on the observability of unobserved states, which is the intrinsic property of the complex system described in Eq.~\eqref{eq:DS_u1u2}. The observability of unobserved states is determined by the dynamics functions $\mathcal{M}_1$ and $\mathcal{M}_2$. For instance, $\mathbf{u}_2$ is fully observable if the observed states $\mathbf{u}_1$ sufficiently dictate the dynamics of $\mathbf{u}_2$ in such a way that observing $\mathbf{u}_1$ over time gives all the necessary information to estimate $\mathbf{u}_2$ without ambiguity. If the $\mathcal{M}_1$ and $\mathcal{M}_2$ are linear functions, the observability matrix condition can be used to determine the full observability. $\mathbf{u}_2$ is not observable if the dynamics function $\mathcal{M}_1$ of the observed state does not include $\mathbf{u}_2$. In this case, the observation of $\mathbf{u}_1$ does not provide any information to infer the unobserved state $\mathbf{u}_2$, and consequently, CGKN will be unable to accurately predict future states using the estimated unobserved states from DA.

Efficient DA can be performed by CGKN with the analytic formulae in Eq.~\eqref{eq:CGKN_CGF}. The computational complexity for this analytic approach is $\mathcal{O}(N^3)$ where $N=d_{\mathbf{v}}$ is the dimension of the latent state. As a comparison, the computational complexity of ensemble DA methods is $\mathcal{O}(J M^2)$, where $J$ is the ensemble size and $M=d_{\mathbf{u}_2}$ is the dimension of unobserved states. To accurately approximate the probabilistic distribution of an $N$-dimensional random vector, an ensemble size $J=c^M$ is typically required, where $c$ denotes the sample size for a single random variable. This leads to a resulting complexity of $\mathcal{O}(c^M M^2)$, which grows exponentially due to the curse of dimensionality. In real-world applications, a much smaller ensemble size is often adopted due to the restriction of computational resources, which inevitably leads to issues originating from sampling errors, and various techniques (e.g., localization) have been developed to empirically deal with insufficient sample size in a high-dimensional space. Therefore, compared with ensemble-based DA methods, which are commonly used for nonlinear models, the analytic DA formulae of the CGKN model can avoid sampling errors and reduce the computational cost. This allows the DA results of CGKN to be incorporated into the loss function during the training procedure. In each training step, the difference between the data of true unobserved state $\mathbf{u}_2^{\star}$ and the posterior mean $\boldsymbol{\mu}$ is quantified by a DA loss term, which promotes the DA performance of the trained CGKN model. More details about the DA loss term and total loss function are summarized in Section~\ref{SSec:CGKN_Learning}.

It is worth noting that the uncertainty quantification of the original unobserved state variables $\mathbf{u}_2$ poses a challenge, which essentially requires mapping the conditional Gaussian distribution of the latent state variables $\mathbf{v}$ through the nonlinear decoder $\boldsymbol{\psi}$. For a weakly nonlinear decoder, this could be achieved efficiently via accessing the Jacobian of the decoder. In cases where the decoder is strongly nonlinear but the dimension of the latent space is relatively low, the unscented transform can serve as an effective method to estimate the uncertainties in $\mathbf{u}_2$. However, the learned decoder of the CGKN model may exhibit strong nonlinearity and operate in a high-dimensional latent space, which prevents an efficient quantification of the uncertainties in $\mathbf{u}_2$ based on existing methods.

To efficiently quantify the uncertainties associated with the posterior mean of the unobserved state $\mathbf{u}_2$, this work explores a post-processing tool based on residual analysis. Residual analysis \citep{anscombe1973graphs, ling1984residuals, chen2021bamcafe} is a method used in statistical modeling and stochastic computing to assess how well a model fits the observed data. With a trained CGKN model, we can perform efficient data assimilation via Eq.~\eqref{eq:CGKN} and the decoder $\boldsymbol{\psi}$ to obtain posterior mean $\{\boldsymbol{\mu}(t_n)\}_{n=0}^{N}$ of unobserved state $\mathbf{u}_2$. The residual $\mathbf{r}$ is then calculated as the absolute difference between true data $\mathbf{u}_2^{\star}$ and the DA posterior mean $\boldsymbol{\mu}$, i.e., $\mathbf{r}(t_n) = |\mathbf{u}_2^{\star}(t_n) - \boldsymbol{\mu}(t_n)|$, for $n = 0,1,..., N$. The residual $\mathbf{r}$ is used to indicate the uncertainties associated with the posterior mean $\boldsymbol{\mu}$. In practice, the residual $\mathbf{r}$ is assumed to be a function of observed state variables $\mathbf{u}_1$, and the function can be approximated by an auxiliary neural network, with the standard regression task based on a dataset of $\{(\mathbf{u}^{\star}_1(t_n), \mathbf{r}(t_n))\}_{n=0}^N$ that are available from a trained CGKN model. Under Gaussian assumption, the output of the trained auxiliary neural network can be regarded as estimated standard deviation of the associated posterior mean from maximum likelihood estimation (MLE). 

\subsection{Learning CGKN from Data}
\label{SSec:CGKN_Learning}

In this work, the proposed CGKN model involves several terms to be determined from data, including the encoder $\boldsymbol{\varphi}$, the decoder $\boldsymbol{\psi}$, and functions $\mathbf{f}_1$, $\mathbf{g}_1$, $\mathbf{f}_2$, and $\mathbf{g}_2$ in the conditional Gaussian nonlinear system of Eq.~\eqref{eq:CGKN}. As illustrated in Fig.~\ref{fig:SchematicDiagram2}, these unknown terms are approximated by an autoencoder that contains both the encoder and the decoder and deep neural networks for the unknown functions in the conditional Gaussian nonlinear system. To train the autoencoder and the deep neural networks jointly, the total loss function consists of four types of loss terms.

The first loss term is the autoencoder loss $L_{\textrm{AE}}$, which essentially ensures that the latent state variables $\mathbf{v}$ contain enough information of the original unobserved state variables $\mathbf{u}_2$. This is achieved by minimizing the difference between the true data $\mathbf{u}_2^{\star}$ and the outputs of autoencoder $\boldsymbol{\psi}(\mathbf{v}^{\star})$, where $\mathbf{v}^{\star}=\boldsymbol{\varphi}(\mathbf{u}^{\star}_2)$. It is worth noting that the tunable nonlinearity within the encoder $\boldsymbol{\varphi}$ allows the conditional linear dynamics of the latent state variables $\mathbf{v}$ in Eq.~\eqref{eq:CGKN}, which is inspired by the Koopman theory, even though the true dynamics of $\mathbf{u}_2$ can be strongly nonlinear. The autoencoder loss $L_{\textrm{AE}}$ is defined as $\mathbb{E}_{\mathbf{u}_2^{\star}}\|\mathbf{u}_2^{\star} - \boldsymbol{\psi}(\boldsymbol{\varphi}(\mathbf{u}_2^{\star}))\|^2.$

The second and third loss terms are the forecast loss of the original state denoted as $L_{\mathbf{u}}$ and the forecast loss of the latent state denoted as $L_{\mathbf{v}}$. The $L_{\mathbf{u}}$ quantifies the difference between true state $\mathbf{u}^{\star}$ and the state prediction $\mathbf{u}$, while the $L_{\mathbf{v}}$ measures the difference between true latent state $\mathbf{v}^{\star} = \boldsymbol{\varphi}(\mathbf{u}_2^{\star})$ and latent state prediction $\mathbf{v}$. The motivation for incorporating the forecast loss of the original state variables $\mathbf{u}$ is straightforward, as one of the applications for CGKN model is to provide state prediction of the true complex dynamical system. On the other hand, we found that the forecast loss $L_{\mathbf{v}}$ of the latent state variables $\mathbf{v}$ can enhance the predictive performance of the CGKN model, and the main reason is that the forecast loss $L_{\mathbf{v}}$ promotes the predictive capability of the conditional Gaussian nonlinear system in Eq.~\eqref{eq:CGKN}. The formulas of forecast loss of the original state $L_{\mathbf{u}}$ and the forecast loss of the latent state $L_{\mathbf{v}}$ are $\mathbb{E}_{\mathbf{u}^{\star}(t_0)}\frac{1}{N_s}\sum_{i=1}^{N_s} \|\mathbf{u}^{\star}(t_i) - \mathbf{u}(t_i)\|^2$ and $\mathbb{E}_{\mathbf{u}^{\star}(t_0)}\frac{1}{N_s}\sum_{i=1}^{N_s} \|\mathbf{v}^{\star}(t_i) - \mathbf{v}(t_i)\|^2$, respectively, where $N_s$ represents the number of forecast steps.

The fourth loss term is the DA loss $L_{\textrm{DA}}$, which is enabled by the efficient analytic formulae in Eq.~\eqref{eq:CGKN_CGF} to obtain the posterior mean $\boldsymbol{\mu}_\mathbf{v}$ of the latent state variables $\mathbf{v}$. It is worth noting that the DA performance of the original unobserved state variables $\mathbf{u}_2$ is of actual interest. In this work, we directly map the posterior mean $\boldsymbol{\mu}_\mathbf{v}$ to approximate the posterior mean $\boldsymbol{\mu}$ of $\mathbf{u}_2$ via the decoder, i.e., $\boldsymbol{\mu} \approx \boldsymbol{\psi}(\boldsymbol{\mu}_\mathbf{v})$. Although such an approximation is often invalid for an arbitrary nonlinear decoder $\boldsymbol{\psi}$, we enforce the difference between the mapped posterior mean $\boldsymbol{\mu}$ and the data of $\mathbf{u}_2$ to be small via the DA loss term, which effectively promotes the nonlinearity within the trained decoder such that the direct mapping of posterior mean values serves as a good approximation. In addition, the DA loss can potentially enhance the discovery of probabilistic relation between the observed state variables $\mathbf{u}_1$ and the unobserved ones $\mathbf{u}_2$ when training the CGKN model. The DA loss $L_{\textrm{DA}}$ is defined as $\mathbb{E}_{\mathbf{u}^{\star}(t_0)}\frac{1}{N_l-N_b}\sum_{\substack{j=N_b+1}}^{N_l} \|\mathbf{u}^{\star}_2(t_j) - \boldsymbol{\mu}(t_j)\|^2$, where $N_l$ is the number of DA steps and $N_b$ is the warm-up period steps.

The total loss function for the training of CGKN model is then defined as:
\begin{equation}
\label{eq:Loss_Target}
\begin{aligned}
    L:= &\lambda_\textrm{AE} L_{\textrm{AE}} + \lambda_{\mathbf{u}} L_{\mathbf{u}} + \lambda_{\mathbf{v}} L_{\mathbf{v}} + \lambda_{\textrm{DA}}L_{\textrm{DA}},
    \end{aligned}
\end{equation}

with
\begin{align}
L_{\textrm{AE}} &:= \mathbb{E}_{\mathbf{u}_2^{\star}}\|\mathbf{u}_2^{\star} - \boldsymbol{\psi}(\boldsymbol{\varphi}(\mathbf{u}_2^{\star}))\|^2,\label{eq:Loss_AE}\\
L_{\mathbf{u}} &:= \mathbb{E}_{\mathbf{u}^{\star}(t_0)}\frac{1}{N_s}\sum_{i=1}^{N_s} \|\mathbf{u}^{\star}(t_i) -  \mathbf{u}(t_i)\|^2,\label{eq:Loss_u}\\
L_{\mathbf{v}} &:= \mathbb{E}_{\mathbf{u}^{\star}(t_0)}\frac{1}{N_s}\sum_{i=1}^{N_s} \|\mathbf{v}^{\star}(t_i) -  \mathbf{v}(t_i)\|^2,\label{eq:Loss_v}\\
L_{\textrm{DA}} &:= \mathbb{E}_{\mathbf{u}^{\star}(t_0)}\frac{1}{N_l-N_b}\sum_{\substack{j=N_b+1}}^{N_l} \|\mathbf{u}^{\star}_2(t_j) - \boldsymbol{\mu}(t_j)\|^2,\label{eq:Loss_DA}
\end{align}

where $\|\cdot\|$ denotes the standard vector $\ell^2$-norm and $\mathbf{u}^{\star}$ indicates data from the true system, and the expectations are estimated based on the samples in the training data. The above loss terms involve some hyper-parameters, including the total time steps $N_s$ for the state forecast, the total time steps $N_l$ for DA, and the warm-up period steps $N_b$ for DA. For the state forecast loss terms, the CGKN model starts from the same initial condition at time $t_0$ as the training data from the true system. For the DA loss term, a warm-up period is often needed due to the effect of an inaccurate initial condition, and the warm-up period is excluded from the evaluation of DA loss term.

The architecture of the CGKN model has been presented in Fig.~\ref{fig:SchematicDiagram2}(a), which consists of three neural-network-based components, including the encoder $\boldsymbol{\varphi}$, the decoder $\boldsymbol{\psi}$, and a neural network $\boldsymbol{\eta}$ to approximate the unknown functions $\mathbf{f}_1$, $\mathbf{g}_1$, $\mathbf{f}_2$, and $\mathbf{g}_2$ in the conditional Gaussian nonlinear system. Learning the unknown coefficients of those neural-network-based components from data is to seek a minimizer of the following optimization problem:
\begin{equation}
\label{eq:opt_problem}
\min_{\boldsymbol{\theta}_{\boldsymbol{\varphi}},\boldsymbol{\theta}_{\boldsymbol{\psi}}, \boldsymbol{\theta}_{\boldsymbol{\eta}}}\Big( \lambda_\textrm{AE} L_{\textrm{AE}}(\boldsymbol{\theta}_{\boldsymbol{\varphi}},\boldsymbol{\theta}_{\boldsymbol{\psi}}) + \lambda_{\mathbf{u}} L_{\mathbf{u}}(\boldsymbol{\theta}_{\boldsymbol{\varphi}},\boldsymbol{\theta}_{\boldsymbol{\psi}}, \boldsymbol{\theta}_{\boldsymbol{\eta}}) + \lambda_{\mathbf{v}} L_{\mathbf{v}}(\boldsymbol{\theta}_{\boldsymbol{\varphi}}, \boldsymbol{\theta}_{\boldsymbol{\eta}}) + \lambda_{\textrm{DA}}L_{\textrm{DA}}(\boldsymbol{\theta}_{\boldsymbol{\psi}}, \boldsymbol{\theta}_{\boldsymbol{\eta}})\Big),
\end{equation}
which can be solved by standard stochastic gradient descent algorithms, and the gradient is readily available via the implementation of the CGKN model on a platform that supports automatic differentiation. A common choice for the gains is $\lambda_{\textrm{AE}} = 1/d_{\mathbf{u}_2}$, $\lambda_{\mathbf{u}} = 1/d_{\mathbf{u}}$, $\lambda_{\mathbf{v}} = 1/d_{\mathbf{u}}$, and $\lambda_{\textrm{DA}} = 1/d_{\mathbf{u}_2}$, which scale each loss function by a factor of the inverse of the respective vector's dimension. The resulting loss functions are the mean squared error (MSE) between the true data and the approximated data. In practice, the gains can also be set as hyper-parameters or manually adjusted based on the domain knowledge of users.

It is important to note that the noise coefficients $\boldsymbol{\sigma}_1$ and $\boldsymbol{\sigma}_2$ of the conditional Gaussian nonlinear system in Eq.~\eqref{eq:CGKN} are required to evaluate the DA loss term for the CGKN model using the analytic formulae in Eq.~\eqref{eq:CGKN_CGF}. The coefficient $\boldsymbol{\sigma}_1$ represents the noise associated with the observed state $\mathbf{u}_1$. Assuming $\boldsymbol{\sigma}_1$ is a diagonal matrix, its diagonal elements can be estimated through the quadratic variation of a pre-trained CGKN model without the DA loss component:
\begin{equation}
    \begin{aligned}
        \mathrm{diag}(\boldsymbol{\sigma}_1) = \sqrt{ \frac{\Delta t}{N_t}\sum_{n=1}^{N_t} \bigl(\dot{\mathbf{u}}^{\star}_1(t_n) -  \dot{\mathbf{u}}_1(t_n)\bigr) \odot \bigl(\dot{\mathbf{u}}_1^{\star}(t_n) -  \dot{\mathbf{u}}_1(t_n)\bigr)},
    \end{aligned}
    \label{eq:sigma_estimation}
\end{equation}
where $\Delta t$ denotes the time step of the training data, $N_t$ represents the total number of time steps in the training data, and the notation $\odot$ signifies element-wise multiplication. In Eq.~\eqref{eq:sigma_estimation}, $\dot{\mathbf{u}}_1^{\star}(t_n)$ is the time derivative of the true observed state variables, which can be computed by taking the numerical derivatives of $\mathbf{u}_1$ at time $t_n$, while $\dot{\mathbf{u}}_1(t_n)$ is the time derivative of the corresponding prediction from the pre-trained CGKN model without the DA loss. The coefficient $\boldsymbol{\sigma}_2$ represents the noise associated with the latent state $\mathbf{v}$, which does not physically exist and, therefore, lacks a true value. It is also assumed to be a diagonal matrix in the form $c\mathbf{I}_{d_{\mathbf{v}}}$, where $c \in \mathbb{R}$  is a scalar and $\mathbf{I}_{d_{\mathbf{v}}} \in \mathbb{R}^{d_{\mathbf{v}} \times d_{\mathbf{v}}}$ is the identity matrix. In addition, $\boldsymbol{\sigma}_2$  in the analytic formulae of Eq.~\eqref{eq:CGKN_CGF} acts as a bias term, and thus $c$ can be treated as either a trainable parameter or a constant. In this work, the scalar $c$ is manually set as a constant.

\section{Numerical Experiments}
\label{Sec:NumericalExperiments}

To demonstrate the capability of CGKN in both state prediction and DA, we test it on several complex dynamical systems and compare its performance against other models. Specifically, the tested systems include the projected stochastic Burgers–Sivashinsky equation, which exhibits strong intermittency and extreme events, the Lorenz 96 system with 40 state variables, and a multiscale non-Gaussian climate phenomenon --- the El Ni\~no Southern Oscillation (ENSO).

The models that have been tested for each system are summarized as follows:
\begin{enumerate}[(i)]
    \item The true model. This corresponds to the governing equation of the complex dynamical systems in Eq.~\eqref{eq:DS_u1u2}. State predictions are made by averaging ensemble simulations from any given initial state, while DA results are obtained using the ensemble Kalman-Bucy filter (EnKBF) \citep{bergemann2012ensemble} based on data from observed states. In this work, the EnKBF is employed exclusively by this model for DA, while other models use the analytic DA formulae from Eq.~\eqref{eq:CGKN_CGF} to ensure a fair comparison. The test results from the true model serve as a benchmark for optimal performance, though they may not always be the best in practical applications.

    \item Conditional Gaussian Koopman network (CGKN). This is the model depicted in Fig.~\ref{fig:SchematicDiagram2}(a). Without requiring any prior knowledge of the governing equations of the true complex dynamical systems, the CGKN model demonstrates performance comparable to the true model in both state prediction and DA. Furthermore, it significantly improves the efficiency of DA due to the analytic DA formulae provided in Eq.~\eqref{eq:CGKN_CGF}.

    \item Standard Koopman network (Standard KoopNet). This model applies the standard Koopman theory to the dynamics of the unobserved state $\mathbf{u}_2$, resulting in a linear governing equation for the latent state variables $\mathbf{v}$, rather than the nonlinear but conditionally linear structure of $\mathbf{v}$ found in the CGKN model. It is worth highlighting that this model still retains a strongly nonlinear structure in the dynamics of $\mathbf{u}_1$. The detailed form of the standard Koopman network is:
    \begin{equation}
        \begin{aligned}
        \label{eq:SimplifiedKoopmanNet}
        &\frac{\diff \mathbf{u}_1}{\diff t} = {\mathbf{f}}_1(\mathbf{u}_1) + {\mathbf{g}}_1(\mathbf{u}_1)\mathbf{v} + {\boldsymbol{\sigma}}_1\dot{\mathbf{W}}_1,\\
        &\frac{\diff \mathbf{v}}{\diff t} = \mathbf{F}_2 + \mathbf{G}_2\mathbf{v} + {\boldsymbol{\sigma}}_2\dot{\mathbf{W}}_2,
        \end{aligned}
    \end{equation}
    where $\mathbf{F}_2 \in \mathbb{R}^{d_{\mathbf{v}}}$ and $\mathbf{G}_2\in \mathbb{R}^{d_{\mathbf{v}} \times d_{\mathbf{v}} }$ are two parameter matrices with other components staying the same as the CGKN model in Eq.~\eqref{eq:CGKN}. This standard application of the Koopman operator remains to impose a conditional Gaussian structure on the model, but the linear governing equation for $\mathbf{v}$ overlooks the contribution of the observed state $\mathbf{u}_1$. Numerical results show that, in comparison to the standard KoopNet model, the CGKN model's conditional linear structure in the governing equation of $\mathbf{v}$ can further enhance its expressiveness. The comparison between the CGKN and the standard KoopNet model highlights the crucial role of the conditional linear, yet still nonlinear, structure in the governing equation of $\mathbf{v}$ in the former. The standard KoopNet can be viewed as a simplified version of CGKN where the latent dynamics do not explicitly depend on the observed states. If the true dynamical system possesses this characteristic, then the expressiveness of the standard KoopNet is sufficient for it to perform comparably to CGKN.
    
    \item Conditional Gaussian regression (CG-Reg). This is a regression model from system identification that involves two steps: (i) identifying significant basis functions from a library of system dynamics using a causal-based metric called causation entropy \citep{almomani2020entropic, fish2021entropic, kim2017causation, almomani2020erfit, elinger2021information, elinger2021causation}, and (ii) estimating the coefficients of these significant basis functions via the least squares method or maximum likelihood estimation. The library is specifically designed to ensure that the resulting regression model conforms to the structure of the conditional Gaussian nonlinear system in Eq.~\eqref{eq:CGKN}, allowing the analytic formulae in Eq.~\eqref{eq:CGKN_CGF} to still be applied for DA \citep{chen2023causality}. In this work, this specially designed library is referred to as the CG-library, which requires that the unobserved state variables $\mathbf{u}_2$ appear linearly in all the basis functions. The detailed form of the CG-Reg model is as follows:
    \begin{equation}
        \begin{aligned}
        \label{eq:KnowledgeReg}
        &\frac{\diff \mathbf{u}_1}{\diff t} = \boldsymbol{\Xi}_1\boldsymbol{\Phi} + {\boldsymbol{\sigma}}_1\dot{\mathbf{W}}_1,\\
        &\frac{\diff \mathbf{u}_2}{\diff t} = \boldsymbol{\Xi}_2\boldsymbol{\Phi} + {\boldsymbol{\sigma}}_2\dot{\mathbf{W}}_2,
        \end{aligned}
    \end{equation}
    where $\boldsymbol{\Phi} = [\phi_1(\mathbf{u}_1, \mathbf{u}_2), \cdots, \phi_M(\mathbf{u}_1, \mathbf{u}_2)]^\mathtt{T} \in \mathbb{R}^{M}$ is a CG-library with $\mathbf{u}_2$ incorporated linearly in each basis function $\phi_i$ while $\boldsymbol{\Xi}_1 \in \mathbb{R}^{d_{\mathbf{u}_1}\times M}$ and $\boldsymbol{\Xi}_2 \in \mathbb{R}^{d_{\mathbf{u}_2}\times M}$ are two matrices storing the coefficients of the basis functions in the CG-Library $\boldsymbol{\Phi}$. Although the CG-Reg model may have better interpretability, it often exhibits worse performance of state prediction and DA compared to the CGKN model, mainly due to the fact that some important basis functions may be missing in the CG-library for real-world applications. This comparison highlights the expressiveness of the neural networks in the CGKN model.

    \item Deep neural network (DNN): It is a black-box surrogate model based on neural networks. It reads:
    \begin{equation}
        \frac{\mathrm{d} \mathbf{u}}{\mathrm{d} t} = \textrm{NN}(\mathbf{u}) + \boldsymbol{\sigma}\dot{\mathbf{W}},
    \end{equation}
    where $\mathbf{u}$ is the original system state variables including both the observed and unobserved ones. This DNN model does not follow the conditional Gaussian structure, and therefore, the analytic formulae in Eq.~\eqref{eq:CGKN_CGF} cannot be applied for efficient DA. Numerical results demonstrate that the DNN model achieves state forecast performance comparable to that of the CGKN model; however, the additional key advantage of the CGKN model lies in its unique efficiency in DA.
\end{enumerate}

For all numerical test cases, we assume that both the training and testing data include observed and unobserved state variables, although the inference of a trained CGKN does not require unobserved state variables. To evaluate the performance of models in both state prediction and DA across all examples, the normalized root mean squared error (NRMSE) is used:
\begin{equation}
    \label{eq:NRMSE}
    \textrm{NRMSE} := \frac{1}{M} \sum_{i=1}^{M} \frac{\sqrt{\frac{1}{N}\sum_{n=1}^N \big( \mathbf{x}_i^{\star}(t_n) - \mathbf{x}_i(t_n) \big)^2}}{\textrm{std}(\mathbf{x}_i^{\star})},
\end{equation}
where $\mathbf{x}^{\star}(t_n) \in \mathbb{R}^{M}$ is the true variables at the time step $t_n$, $\mathbf{x}(t_n)$ is the approximated variables, $\mathbf{x}_i(t_n)$ denotes the $i$-th variable, and $\textrm{std}(\cdot)$ corresponds to the standard deviation estimated by the samples from different time steps $t_n=1,...,N$. The NRMSE between true state variables and predictive ones is referred to as the forecast error, and the NRMSE between the true state variables and estimated state variables from DA is referred to as the DA error. Both forecast error and DA error are calculated using test data. The forecast error is calculated based on the same total forecast time steps in training. In contrast, the DA error is calculated over the whole time range of the test dataset. 

\subsection{The projected stochastic Burgers–Sivashinsky equation: A strongly nonlinear system with intermittency and extreme events}

\begin{figure}[H]
    \centering
    \includegraphics[width=1\linewidth]{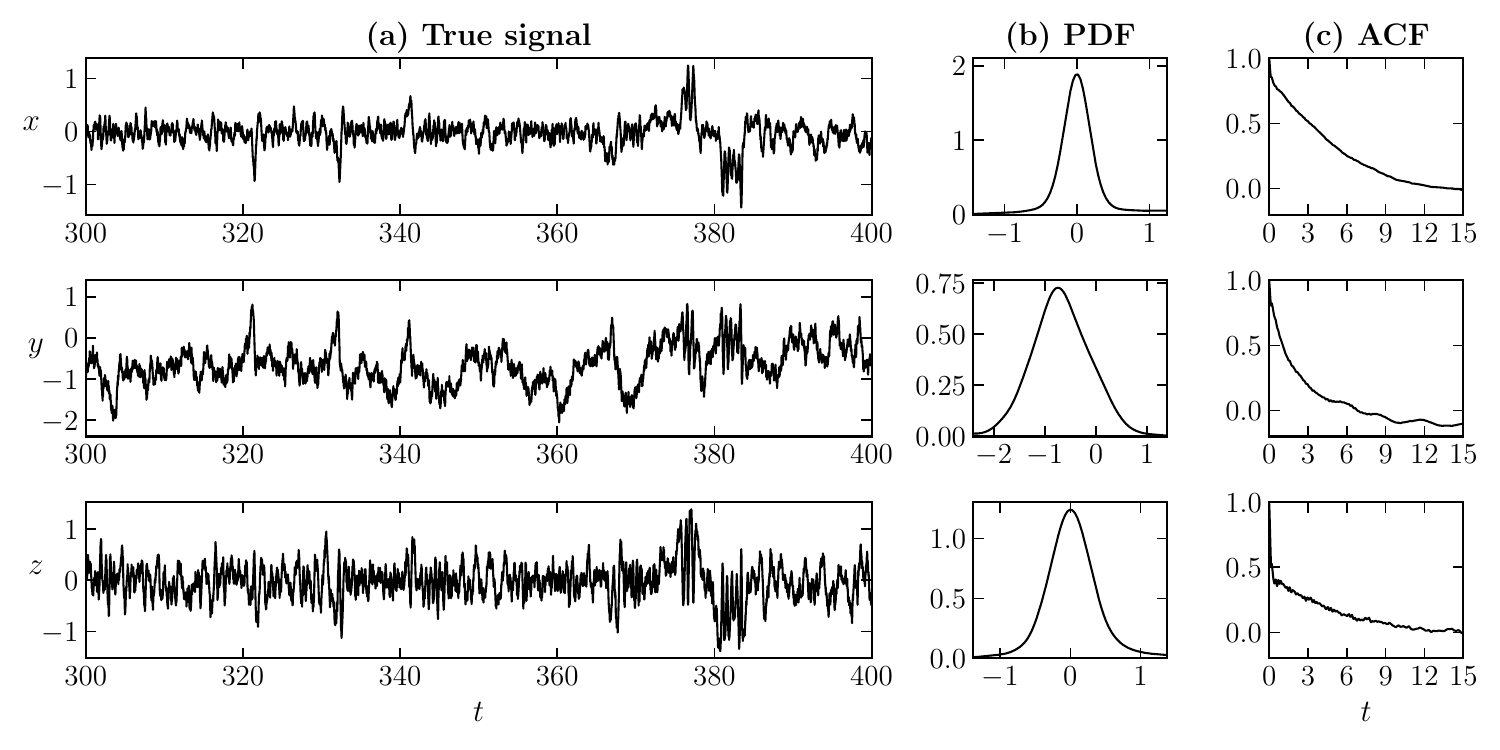}
    \caption{Simulation results of the projected stochastic Burgers–Sivashinsky equation, with (a) time series of each state variable, (b) the PDFs of the corresponding state variable, and (c) the ACFs of the corresponding state variable. It should be noted that the PDFs and ACFs are estimated from much longer simulations than the one presented in panel (a).}
    \label{fig:PSBSE_Property}
\end{figure}

The Fourier-Galerkin projection of the stochastic Burgers–Sivashinsky equation
\begin{equation*}
\partial_t u  = \big( \nu \partial_{xx} u  + \lambda u  - u  \partial_x u\big) + \dot{W}(t,x)
\end{equation*} subject to homogeneous Dirichlet boundary conditions
is the so-called projected stochastic Burgers–Sivashinsky equation (PSBSE) \citep{chekroun2014stochastic}:
\begin{equation}
    \begin{aligned}
    \label{eq:PSBSE}
        \frac{\diff x}{\diff t} &= \beta_xx + \alpha xy + \alpha yz + \sigma_x\dot{W}_x,\\
        \frac{\diff y}{\diff t} &= \beta_yy - \alpha x^2 + 2\alpha xz + \sigma_y\dot{W}_y,\\
        \frac{\diff z}{\diff t} &= \beta_zz - 3\alpha xy + \sigma_z\dot{W}_z.
    \end{aligned}
\end{equation}
The PSBSE in Eq.~\eqref{eq:PSBSE} is a stochastic differential equation with energy-conserving quadratic terms and additive Gaussian white noise. The coefficient of the linear terms $\beta_x$ is positive to introduce instability into the system, and $\beta_y$ and $\beta_z$ are negative to linearly damp the system. The coefficient $\alpha > 0$ controls the non-linearity of the system. In this example, the simulation parameters are set as:
\begin{equation}
\begin{gathered}
  \beta_x=0.2, \quad \beta_y=-0.3, \quad \beta_z=-0.5, \quad \alpha = 5, \\
  \quad \sigma_x=0.3, \quad \sigma_y=0.5, \quad  \sigma_z=0.5, \\
  \Delta t = 0.001, \quad \mbox{and} \quad [x_0, y_0, z_0]^{\mathtt{T}} = [1,1,1]^\mathtt{T}.
\end{gathered}
\end{equation}
The simulation results of 1000 time units are generated and provide the data with a sub-sampling step size $\Delta t=0.01$ time unit, where the first 800 units are utilized for training and the remaining 200 units are employed for testing. Figure~\ref{fig:PSBSE_Property} shows part of the simulation of PSBSE in panel (a), and the probability density functions (PDFs) and auto-correlation functions (ACFs) in panels (b) and (c), respectively. The selection of parameters plays a crucial role in the system's behavior: the coefficient of linear terms $\beta_x = 0.2$ introduces instability, while $\beta_y = -0.3$ and $\beta_y = -0.5$ linearly dampens the system. Additionally, setting $\alpha = 5$ contributes significant non-linearity to the system. Under the parameter settings, the true system demonstrates several complex characteristics, including intermittency, extreme events, and non-Gaussian statistics, as illustrated in Figure~3.1. Among the three states, $x$ is set as the observed state, while $y$ and $z$ are set as unobserved states in this test case. This setup of observed and unobserved states aligns with real-world applications since the state $x$ is the largest scale variable in the system. Additional test results of different setups of observed and unobserved states can be found in \ref{Sec:PSBSE_Extra}.

With the encoder $\boldsymbol{\varphi}$, the decoder $\boldsymbol{\psi}$ and the neural network $\boldsymbol{\eta}(x)$ that outputs ${f}_1(x)$, $\mathbf{g}_1(x)$, $\mathbf{f}_2(x)$, and $\mathbf{g}_2(x)$, the CGKN model for this example is written as:
\begin{equation}
    \label{eq:PSBSE_CGKN}
    \begin{aligned}
        \frac{\diff x}{\diff t} &= {{f}}_1(x) + {\mathbf{g}}_1(x)\mathbf{v} + {\sigma}_x\dot{W}_x, \\
        \frac{\diff \mathbf{v}}{\diff t} &= {\mathbf{f}}_2(x) + {\mathbf{g}}_2(x)\mathbf{v} + {\boldsymbol{\sigma}}_{\boldsymbol{v}}\dot{\mathbf{W}}_{\boldsymbol{v}},
    \end{aligned}
\end{equation}
where the latent state variables $\mathbf{v}$ are transformed from unobserved state variables $[y,z]^\mathtt{T}$ via the encoder $\boldsymbol{\varphi}$. The dimension of $\mathbf{v}$ is set as 10 in this test case, and the sub-networks in the CGKN model are $\boldsymbol{\varphi}: \mathbb{R}^{2} \mapsto \mathbb{R}^{10}$, $\boldsymbol{\psi}: \mathbb{R}^{10} \mapsto \mathbb{R}^{2}$, and $\boldsymbol{\eta}: \mathbb{R}^{1} \mapsto \mathbb{R}^{121}$ with number of parameters $\boldsymbol{\theta_{\boldsymbol{\varphi}}}$, $\boldsymbol{\theta_{\boldsymbol{\psi}}}$, and $\boldsymbol{\theta_{\boldsymbol{\eta}}}$ be 8778, 8770, and 5785 respectively. The fully connected neural networks are used in CGKN, with the encoder $\boldsymbol{\varphi}$, decoder $\boldsymbol{\psi}$ and neural networks $\boldsymbol{\eta}$ each consisting of five layers. The ReLU activation function is applied to all networks in CGKN. The training settings of the CGKN include: state prediction time range $[t_0,t_{N_s}]$ is $0.1$ time units, and the DA time range $[t_0,t_{N_l}]$ is $50$ time units with the warm-up period set as $t_{N_b} = 3$ time units. The weights in the target loss function are set as $\lambda_{\textrm{AE}} = 1/d_{\mathbf{u}_2}$, $\lambda_{\mathbf{u}} = \lambda_{\mathbf{v}} = 1/d_{\mathbf{u}}$, and $\lambda_{\textrm{DA}} = 1/d_{\mathbf{u}_2}$.  The gradient descent optimizer is set as Adam with an initial learning rate of 1e-3 and a cosine annealing scheduler.

\begin{table}[H]
\caption{Test results of state forecast and data assimilation of all models for the example of projected stochastic Burgers–Sivashinsky equation. The errors are the normalized root mean squared error (NRMSE) between true values and approximated values.}
\label{tab:PSBSE_Test_Errors}
\begin{adjustbox}{max width=1.\textwidth,center}
\begin{tabular}{|c|c|c|c|c|c|c|}
\hline
\diagbox{Models}{Errors} & Forecast Error & DA Error \\
\hline
True Model & 2.4566e-01 & 6.9466e-01 \\
\hline
CGKN & 2.8389e-01 & 7.2776e-01 \\
\hline
Standard KoopNet & 3.7600e-01 & 7.5069e-01\\
\hline
CG-Reg & 2.7766e-01 & 1.2884e+00 \\
\hline
DNN & 2.4728e-01  & ---  \\
\hline
\end{tabular}
\end{adjustbox}
\end{table}

We investigate the performance of both state prediction and DA of all models on the test data and report the results in Table~\ref{tab:PSBSE_Test_Errors}. The forecast error is calculated by 0.1 time units as the forecast time range and the DA Error is based on 200 time units as the DA time range. For the true model, the ensemble number is set to 100 for both the ensemble mean forecast and the EnKBF. The error in the true model primarily arises from intrinsic random noise in the system, along with sampling errors of the ensemble methods. Consequently, the test results of the true model serve as a benchmark for evaluating the performance of other models. Among all other models, the CGKN model provides the best overall performance in both state forecast and DA. Although the DNN model slightly outperforms the CGKN model for state forecast, its black-box nonlinear architecture demands more computationally expensive DA techniques, while the CGKN model can exploit the analytic formulae of Eq.~\eqref{eq:CGKN_CGF} for efficient DA.
The forecast error from the CG-Reg model is slightly better than that of the CGKN model. The main reason is that, except for the nonlinear $yz$ term, all other terms in the true system of Eq.~\eqref{eq:PSBSE} can still be successfully identified by the CG-Reg model if all those terms have been included in the CG library. On the other hand, the CG-Reg model provides the worst performance of DA, indicating that the nonlinear term of $yz$ in the governing equation of $x$ plays an important role in the probabilistic relation between the observed state variable $x$ and the unobserved ones $y$ and $z$. Regarding the standard KoopNet model, the state forecast error is expected to be larger than that of the CGKN model, primarily due to the omission of the contribution from the observed state variable $x$ in the governing equations of the unobserved state variables.

\begin{figure}[H]
    \centering
    \includegraphics[width=\linewidth]{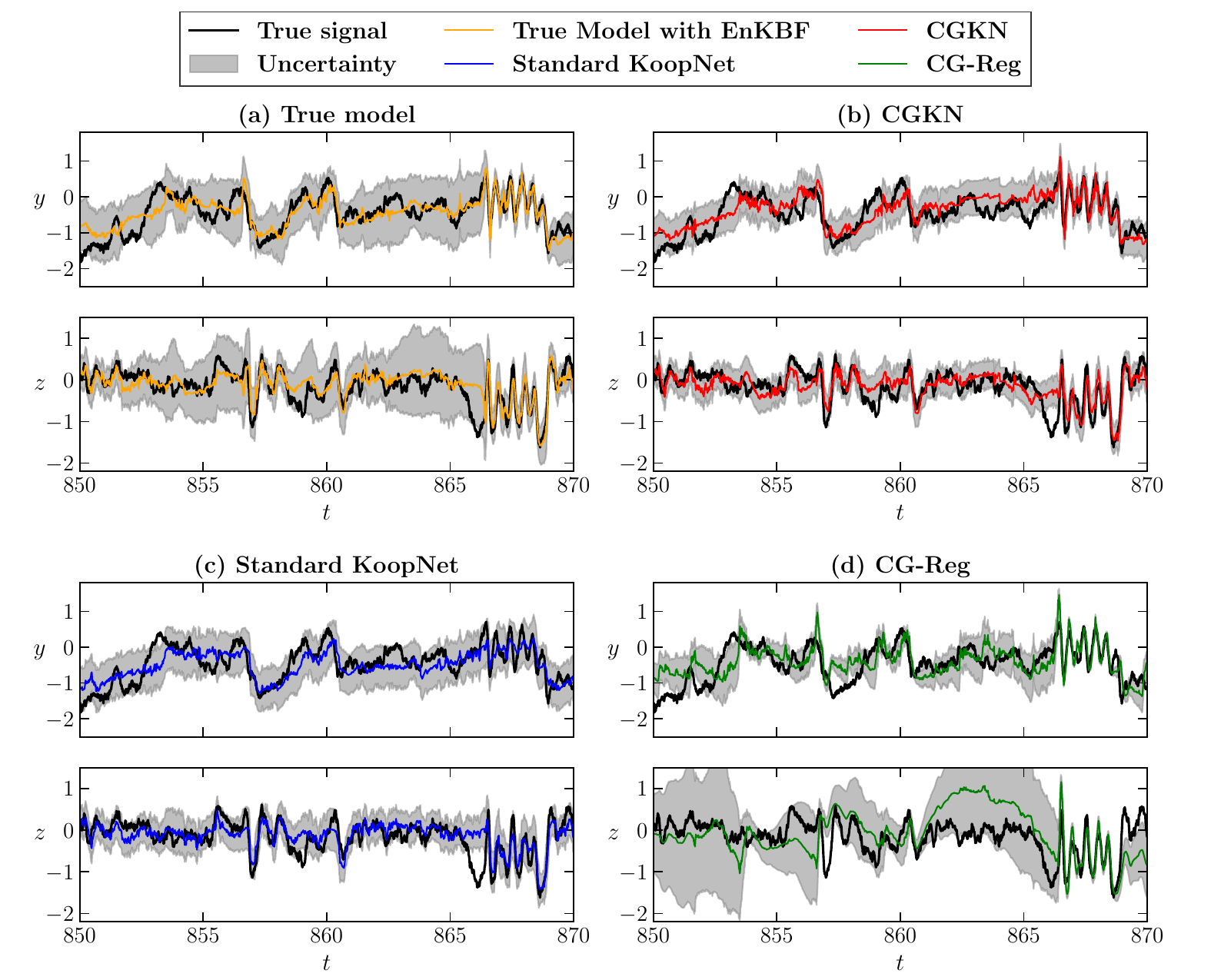}
    \caption{DA results of the projected stochastic Burgers–Sivashinsky equation. EnKBF is used for the true model and analytic formulae are used for the other three models. The uncertainties are indicated by the grey-colored regions, which correspond to two estimated standard deviations from the posterior mean. The uncertainty area of the CGKN model and the standard KoopNet model are estimated based on the method of residual analysis. }
    \label{fig:PSBSE_DA}
\end{figure}

Figure~\ref{fig:PSBSE_DA} presents the DA results from the four models for estimating the unobserved states $[y, z]^\mathtt{T}$ based on the observed state $x$. To obtain the posterior mean estimates, the EnKBF is employed for the true model, while the analytic formulae in Eq.~\eqref{eq:CGKN_CGF} are used for the other three models. The uncertainty regions for the true model and the CG-Reg model are derived from the EnKBF and the covariance evolution formula in Eq.~\eqref{eq:CGKN_CGF}, respectively. In contrast, the uncertainty regions for the CGKN and the standard Koopman network are obtained using the residual analysis method described in Section~\ref{SSec:CGKN_DA}. The comparable performance of the posterior mean estimation to that of the true model underscores the effectiveness of the CGKN in DA. Furthermore, the estimated uncertainty region encompasses most of the results from the true model, even though it remains narrower than that produced by the true model with EnKBF. In this numerical test case, the DA results from the standard KoopNet model shown in Fig.~\ref{fig:PSBSE_DA}(c) are comparable to those of the CGKN model, while the results from the CG-Reg model in Fig.~\ref{fig:PSBSE_DA}(d) are noticeably inferior to those of the other models, particularly for the unobserved state variable $z$.

\subsection{Lorenz 96 system: A forty-dimensional chaotic system}

\begin{figure}[H]
    \centering
    \includegraphics[width=0.7\linewidth]{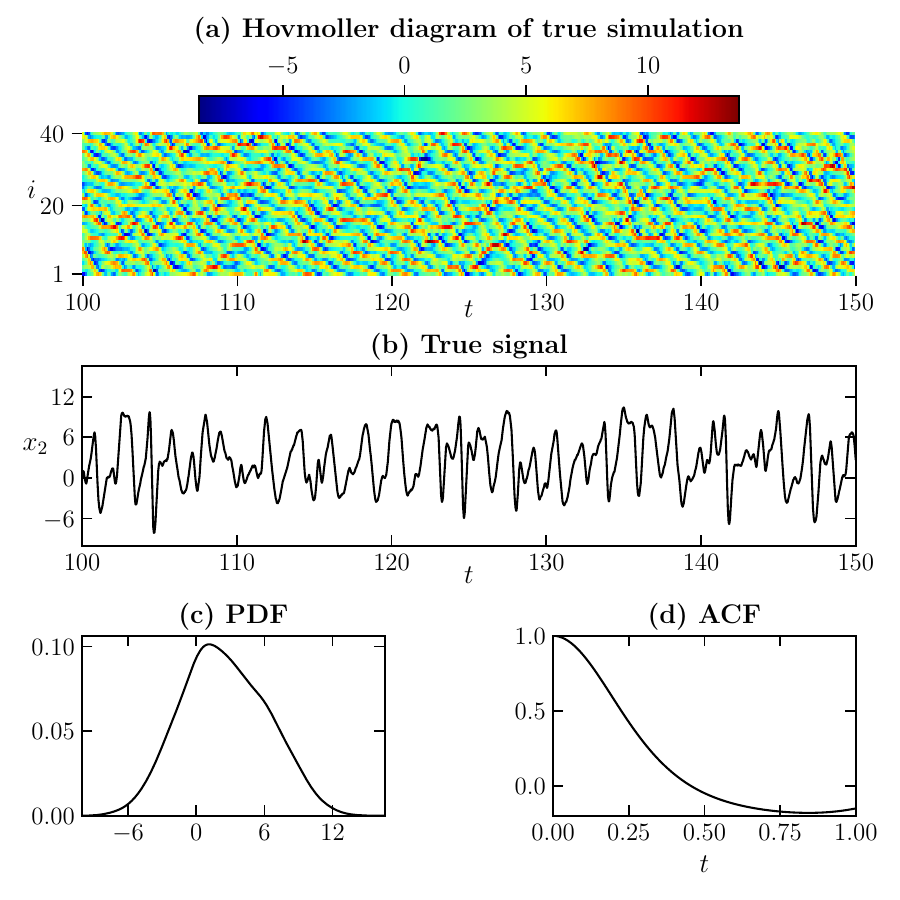}
    \caption{Simulation results and statistics of the Lorenz 96 system, with (a) the Hovmoller diagram of the spatiotemporal patterns, (b) time series of $x_2$, (c) the PDF of $x_2$, and (d) the ACF of $x_2$. Note that the PDF and ACF are estimated from a simulation longer than the one presented in panel (b). The behavior of other state variables is similar to $x_2$ since the system is statistically homogeneous.}
    \label{fig:L96(40)_Property}
\end{figure}

As a widely used numerical example for DA and climate modeling, the governing equation of the single-scale Lorenz 96 system with stochastic noise is as follows \citep{lorenz1996predictability, wilks2005effects, karimi2010extensive}:
\begin{equation}
\begin{aligned}
\label{eq:Lorenz96}
        \frac{\diff x_i}{\diff t}=(x_{i+1} - x_{i-2})x_{i-1} - x_i +F + \sigma\dot{W}_i, \quad i=1,2,3 \cdots, I,
    \end{aligned}
\end{equation}
where $I = 40$, $F = 8$ and $\sigma = 0.5$. Under the specified parameter settings, the Lorenz 96 system exhibits chaotic and wave-like behaviors, allowing it to be regarded as a discretization of atmospheric flow on a latitude circle. Therefore, the current choice of parameters makes the system particularly useful for studying weather patterns and atmospheric dynamics in a controlled yet realistic setting. We simulate the system for $500$ time units with the time step size $\Delta t=10^{-3}$ with the above parameters. With a sub-sampling time step size $\Delta t=0.01$, the first $300$ time units are used as training data, and the rest $200$ time units are used for testing. Part of the spatial-temporal simulation results of the true system, including the time series, the PDF, and the ACF of the state variable $x_2$, are shown in Fig.~\ref{fig:L96(40)_Property}. Since the state variables in this Lorenz 96 system are statistically homogeneous, i.e., each state variable possesses identical statistical properties, the results of other state variables closely resemble the ones of $x_2$ shown in Fig.~\ref{fig:L96(40)_Property}. The PDF and ACF are calculated based on all the 300 time units of training data. In addition to the statistical homogeneity, locality accounts for another feature of the Lorenz 96 system, i.e., the dynamics of a state variable only depend on its nearby state variables. In this numerical test case, a classic DA setup is employed where every second state of the Lorenz 96 system was designated unobserved. More specifically, the system is partially observed with the observed state variables being odd indexed, i.e., $\mathbf{u}_1 = [x_1, x_3, \cdots, x_{39}]^\mathtt{T}$, and the unobserved ones being even indexed, i.e., $\mathbf{u}_2 = [x_2, x_4, \cdots, x_{40}]^\mathtt{T}$. From this setup, we demonstrate that the physics information, including homogeneous property and local spatial structure, can be compiled into CGKN, which forces the model to be consistent with prior conditions and reduces computational costs.

In this work, locality and homogeneity are both incorporated into the models if possible. More specifically, we assume that $\diff x_i / \diff t$ only depends on the local state variable $x_i$ and two nearby state variables $[x_{i-2}, x_{i-1}, x_{i+1}, x_{i+2}]^\mathtt{T}$. On the other hand, the homogeneity indicates that the model that approximates the dynamics $\diff x_i / \diff t$ based on its dependent state variables is consistent for every $i$. It is worth noting that the models with conditional Gaussian structure only allow incorporating the homogeneity of the model form for the observed state variables and the unobserved ones separately, while the training data can still inform the models that all state variables of the true system are statistically homogeneous.

In the CGKN model, the autoencoder is also designed to account for locality and homogeneity. For instance, the unobserved state variables $x_{2i-2}$ and $x_{2i+1}$ nearby the odd-indexed observed state variable $x_{2i-1}$ are transformed to latent state $\mathbf{v}^{(i)} = [\mathbf{v}_1^{(i)}, \mathbf{v}_2^{(i)}, \cdots, \mathbf{v}_J^{(i)}]$ via encoder $\boldsymbol{\varphi}$ for $i=1, 2, \cdots, 20$.

With the encoder $\boldsymbol{\varphi}$, decoder $\boldsymbol{\psi}$ and the DNNs $\boldsymbol{\eta}$ including $\textrm{NN}^{(\mathbf{v})} = [\textrm{NN}^{(\mathbf{v}_1)}, \textrm{NN}^{(\mathbf{v}_2)}, \cdots, \textrm{NN}^{(\mathbf{v}_J)}]$ and $\textrm{NN}^{(\mathbf{u}_1)}$, the framework of CGKN with locality and homogeneity reads:

\begin{equation}
    \begin{aligned}
    \frac{\mathrm{d} x_{2i-1}}{\mathrm{d} t} =\, &\textrm{NN}^{(\mathbf{u}_1)}_0 +
    \\ &\textrm{NN}^{(\mathbf{u}_1)}_1 \mathbf{v}_1^{(i-1)} + \textrm{NN}^{(\mathbf{u}_1)}_2 \mathbf{v}_2^{(i-1)} + \cdots +  \textrm{NN}^{(\mathbf{u}_1)}_{J} \mathbf{v}_J^{(i-1)} +
    \\ &\textrm{NN}^{(\mathbf{u}_1)}_{J+1} \mathbf{v}_1^{(i)} + \textrm{NN}^{(\mathbf{u}_1)}_{J+2} \mathbf{v}_2^{(i)} + \cdots +  \textrm{NN}^{(\mathbf{u}_1)}_{2J} \mathbf{v}_J^{(i)} +
    \\ &\textrm{NN}^{(\mathbf{u}_1)}_{2J+1} \mathbf{v}_1^{(i+1)} + \textrm{NN}^{(\mathbf{u}_1)}_{2J+2} \mathbf{v}_2^{(i+1)} + \cdots +  \textrm{NN}^{(\mathbf{u}_1)}_{3J} \mathbf{v}_J^{(i+1)} +
    \\ & {\sigma}_x\dot{W}_x
    \\ \frac{\mathrm{d} \mathbf{v}^{(i)}_j}{\mathrm{d} t} =\, &\textrm{NN}^{(\mathbf{v}_j)}_0 +
    \\ &\textrm{NN}^{(\mathbf{v}_j)}_{1}\mathbf{v}_1^{(i-2)} + \textrm{NN}^{(\mathbf{v}_j)}_{2}\mathbf{v}_2^{(i-2)} + \cdots + \textrm{NN}^{(\mathbf{v}_j)}_{J}\mathbf{v}_J^{(i-2)} +
    \\ &\textrm{NN}^{(\mathbf{v}_j)}_{J+1}\mathbf{v}_1^{(i-1)} + \textrm{NN}^{(\mathbf{v}_j)}_{J+2}\mathbf{v}_2^{(i-1)} + \cdots + \textrm{NN}^{(\mathbf{v}_j)}_{2J}\mathbf{v}_J^{(i-1)} +
    \\ &\textrm{NN}^{(\mathbf{v}_j)}_{2J+1}\mathbf{v}_1^{(i)} + \textrm{NN}^{(\mathbf{v}_j)}_{2J+2}\mathbf{v}_2^{(i)} + \cdots + \textrm{NN}^{(\mathbf{v}_j)}_{3J}\mathbf{v}_J^{(i)} +
    \\ &\textrm{NN}^{(\mathbf{v}_j)}_{3J+1}\mathbf{v}_1^{(i+1)} + \textrm{NN}^{(\mathbf{v}_j)}_{3J+2}\mathbf{v}_2^{(i+1)} + \cdots + \textrm{NN}^{(\mathbf{v}_j)}_{4J}\mathbf{v}_J^{(i+1)} +
    \\ &\textrm{NN}^{(\mathbf{v}_j)}_{4J+1}\mathbf{v}_1^{(i+2)} + \textrm{NN}^{(\mathbf{v}_j)}_{4J+2}\mathbf{v}_2^{(i+2)} + \cdots + \textrm{NN}^{(\mathbf{v}_j)}_{5J}\mathbf{v}_J^{(i+2)} +
    \\ & {\sigma}_{\mathbf{v}}\dot{W}_{\mathbf{v}}\\
    i =\, & 1, \cdots, 20;\, j= 1, 2, \cdots, J.
    \end{aligned}
\end{equation}

In this test case, the parameter $J$ is selected as 6, and therefore, the dimension of latent state $d_{\mathbf{v}}$ in Eq.\eqref{eq:CGKN} is $20J=120$. The $\textrm{NN}_i^{(\mathbf{u}_1)}$ stands for the $(i+1)$-th output of the network $\textrm{NN}^{(\mathbf{u}_1)} \in \mathbb{R}^{3} \mapsto \mathbb{R}^{19}$ which takes input $(x_{2i-3}, x_{2i-1}, x_{2i+1})$ for dynamics $\mathrm{d} x_{2i-1} / \mathrm{d} t$ and similar to $\textrm{NN}^{(\mathbf{v}_j)} \in \mathbb{R}^{3} \mapsto \mathbb{R}^{31}$ which takes input $(x_{2i-3}, x_{2i-1}, x_{2i+1})$ for dynamics $\mathrm{d} \mathbf{v}^{(i)}_j / \mathrm{d} t$. Since all the neural networks $\textrm{NN}^{(\mathbf{v}_j)}$ take the same inputs for each $j=1, 2, \cdots, J$, they are merged into a single neural network $\textrm{NN}^{(\mathbf{v})}:\in \mathbb{R}^{3} \mapsto \mathbb{R}^{186}$ which takes input $(x_{2i-3}, x_{2i-1}, x_{2i+1})$ for dynamics $\mathrm{d} \mathbf{v}^{(i)} / \mathrm{d} t = [\mathrm{d} \mathbf{v}_1^{(i)} / \mathrm{d} t, \mathrm{d} \mathbf{v}_2^{(i)} / \mathrm{d} t, \cdots, \mathrm{d} \mathbf{v}_J^{(i)} / \mathrm{d} t]$. When iterating $\textrm{NN}^{(\mathbf{u}_1)}$ across all dynamics of observed state, the outputs will formalize the $\mathbf{f}_1 \in \mathbb{R}^{20 \times 1}$ and $\mathbf{g}_1 \in \mathbb{R}^{20 \times 120}$ in Eq.~\eqref{eq:CGKN}. Similarly, iterating $\textrm{NN}^{(\mathbf{v})}$ across all dynamics of latent state will formalize the $\mathbf{f}_2 \in \mathbb{R}^{120\times 1}$ and $\mathbf{g}_2 \in \mathbb{R}^{120\times 120}$. One thing should be noted is that, due to the locality strategy, the $\mathbf{f}_2$ and $\mathbf{g}_2$ are sparse matrices with zero elements representing non-interaction with the uncorrelated states and non-zero elements filled by the outputs of $\textrm{NN}^{(\mathbf{u}_1)}$ and $\textrm{NN}^{(\mathbf{v})}$ respectively. The sub-networks in the CGKN with locality and homogeneity of this test case are $\boldsymbol{\varphi}: \mathbb{R}^{2} \mapsto \mathbb{R}^{6}$, $\boldsymbol{\psi}: \mathbb{R}^{6} \mapsto \mathbb{R}^{2}$ and $\boldsymbol{\eta} = [\textrm{NN}^{(\mathbf{u}_1)}, \textrm{NN}^{(\mathbf{v})}]$ with the number of parameters $\boldsymbol{\theta}_{\boldsymbol{\varphi}}$, $\boldsymbol{\theta}_{\boldsymbol{\psi}}$ and $\boldsymbol{\theta}_{\boldsymbol{\eta}}$ be 8646, 8942 and 5757, respectively. In the CGKN, the encoder $\boldsymbol{\varphi}$, decoder $\boldsymbol{\psi}$ and sub-networks $\boldsymbol{\eta}$ including $\textrm{NN}^{(\mathbf{v})}$ and  $\textrm{NN}^{(\mathbf{u}_1)}$ are all fully connected networks with 5, 5, 4, and 4 layers, respectively. The activation function is chosen as ReLU. The training setup for CGKN is as follows: the state forecast horizon is $t_{N_s} = 0.2$ time units, the DA horizon is $t_{N_l} = 60$ time units with the first $t_{N_b} = 2$ time units be cut out. The weights in the target loss function are set as $\lambda_{\textrm{AE}} = 1/d_{\mathbf{u}_2}$, $\lambda_{\mathbf{u}} = \lambda_{\mathbf{v}} = 1/d_{\mathbf{u}}$, and $\lambda_{\textrm{DA}} = 1/d_{\mathbf{u}_2}$. The gradient descent optimizer is selected as Adam with an initial learning rate of 1e-3 and a cosine annealing scheduler. It is worth noting that the complexity of the CGKN incorporated with the locality and homogeneity does not increase with the dimensionality of the Lorenz 96 system which makes CGKN scalable to even higher dimensions.

\begin{table}[H]
\caption{Test results of state forecast and data assimilation of all models for the example of Lorenz 96 system. The errors are the normalized root mean squared error (NRMSE) between true values and approximated values.}
\label{tab:L96_Test_Errors}
\begin{adjustbox}{max width=1.\textwidth,center}
\begin{tabular}{|c|c|c|c|c|c|c|}
\hline
\diagbox{Models}{Errors} & Forecast Error & DA Error \\
\hline
True Model & 4.3874e-02 & 6.9582e-02 \\
\hline
CGKN & 6.3738e-02 & 9.4883e-02   \\
\hline
Standard KoopNet & 3.0589e-01 & 2.7252e-0 \\
\hline
CG-Reg & 2.7631e-01 & 3.5670e-01\\
\hline
DNN & 4.5174e-02  & ---  \\
\hline
\end{tabular}
\end{adjustbox}
\end{table}

The test results for the different models are reported in Table~\ref{tab:L96_Test_Errors}. For the true model, we use the mean of ensemble simulations for state prediction and employ the EnKBF for DA, with an ensemble size of 100 for both tasks. Among all the other models, the CGKN is the only one that achieves similar performance in both state forecast and DA compared to the true model. In terms of state forecast, the DNN model performs slightly better than the CGKN model. However, the CGKN's advantage lies in its ability to avoid the computationally expensive DA techniques typically required for nonlinear models, making it significantly more efficient for DA. Regarding models that allow efficient DA, the standard KoopNet model slightly outperforms the CG-Reg model; however, both models underperform compared to the CGKN model. 

\begin{figure}[H]
    \centering
    \includegraphics[width=0.7\linewidth]{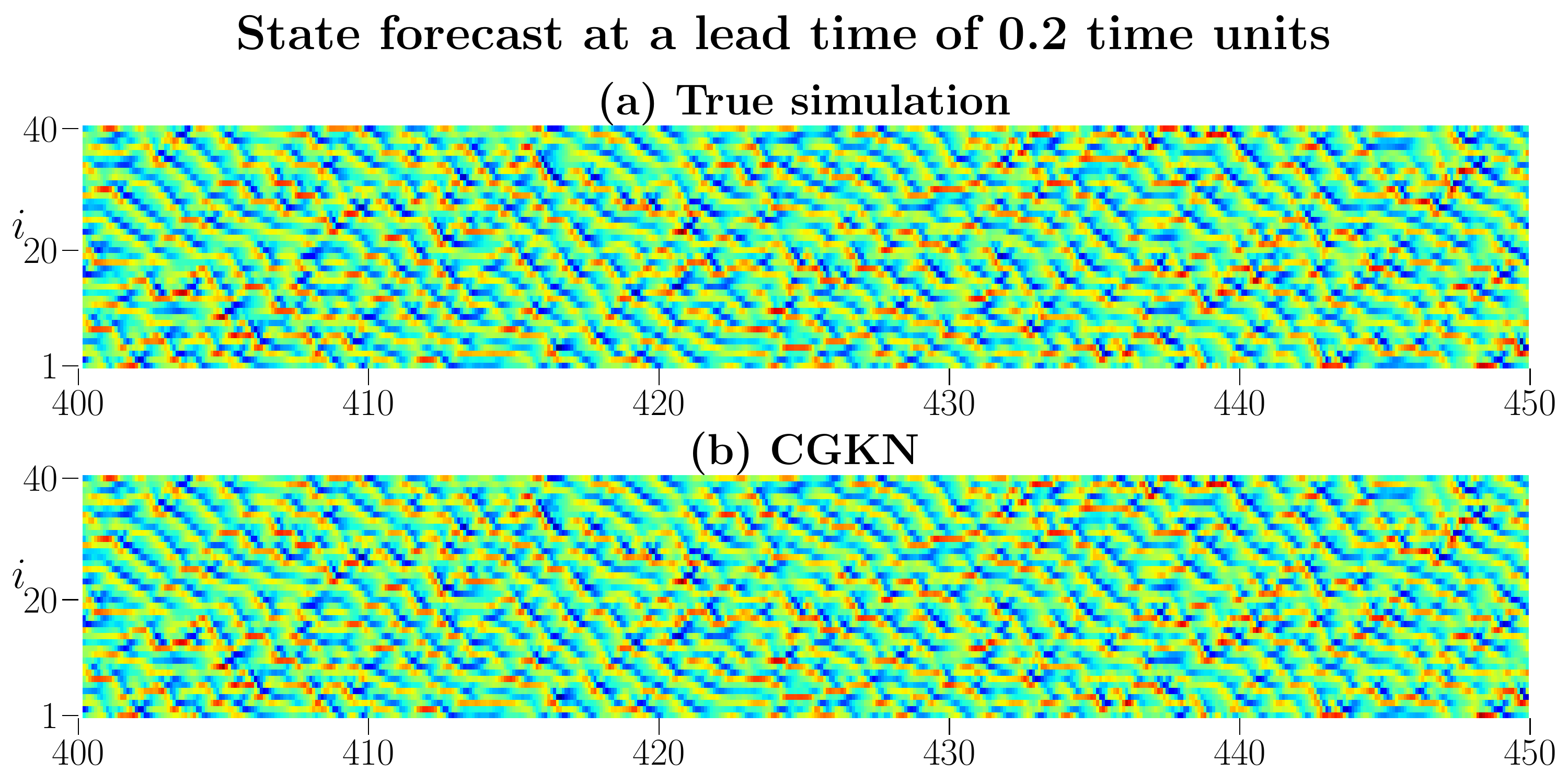}
    \includegraphics[width=0.7\linewidth]{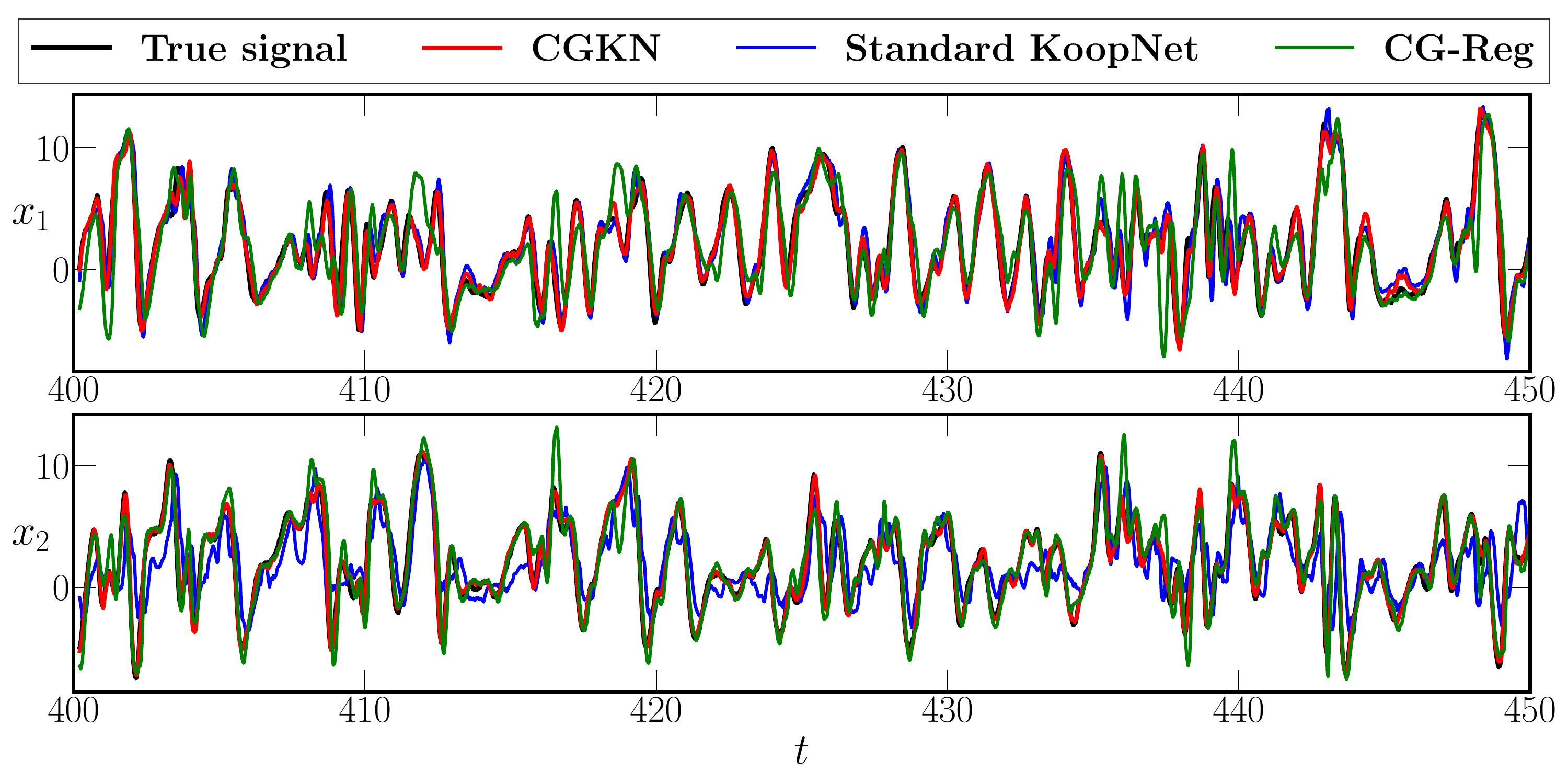}
    \caption{State forecast results with a lead time of 0.2 time units. Panel (a) and (b) are the Hovmoller diagram of the true system and results from the CGKN model. The bottom two panels show the true signal of states $x_1$ and $x_2$ and the corresponding results from three different models.}
    \label{fig:L96_SF}
\end{figure}

Figure~\ref{fig:L96_SF} illustrates the true state variables alongside the prediction results from different models with a lead time of 0.2 time units. The Hovmoller diagram (namely, the spatiotemporal field) of the true system and the simulation results from the CGKN model indicate that the CGKN successfully captures the overall chaotic behavior and wave patterns. The time series of state variables $x_1$ and $x_2$ from the true system, along with the forecast results from the various models, are also presented in Fig.~\ref{fig:L96_SF}. The results for other observed and unobserved state variables are similar to those of $x_1$ and $x_2$ due to the statistical homogeneity of this test case and are thus omitted. Notably, both the standard KoopNet model and the CG-Reg model exhibit discrepancies compared to the truth, while the CGKN model shows results that are more accurate. Although the standard KoopNet model performs reasonably well in forecasting the observed variable $x_1$, its performance for the unobserved state variable $x_2$ is inferior to that of the CGKN model. This discrepancy primarily arises from the neglect of the contribution of observed state variables to the dynamics of unobserved state variables in the standard KoopNet model, which is a crucial factor in the true system due to the quadratic term in Eq.~\eqref{eq:Lorenz96}.

\begin{figure}[H]
    \centering
    \includegraphics[width=\linewidth]{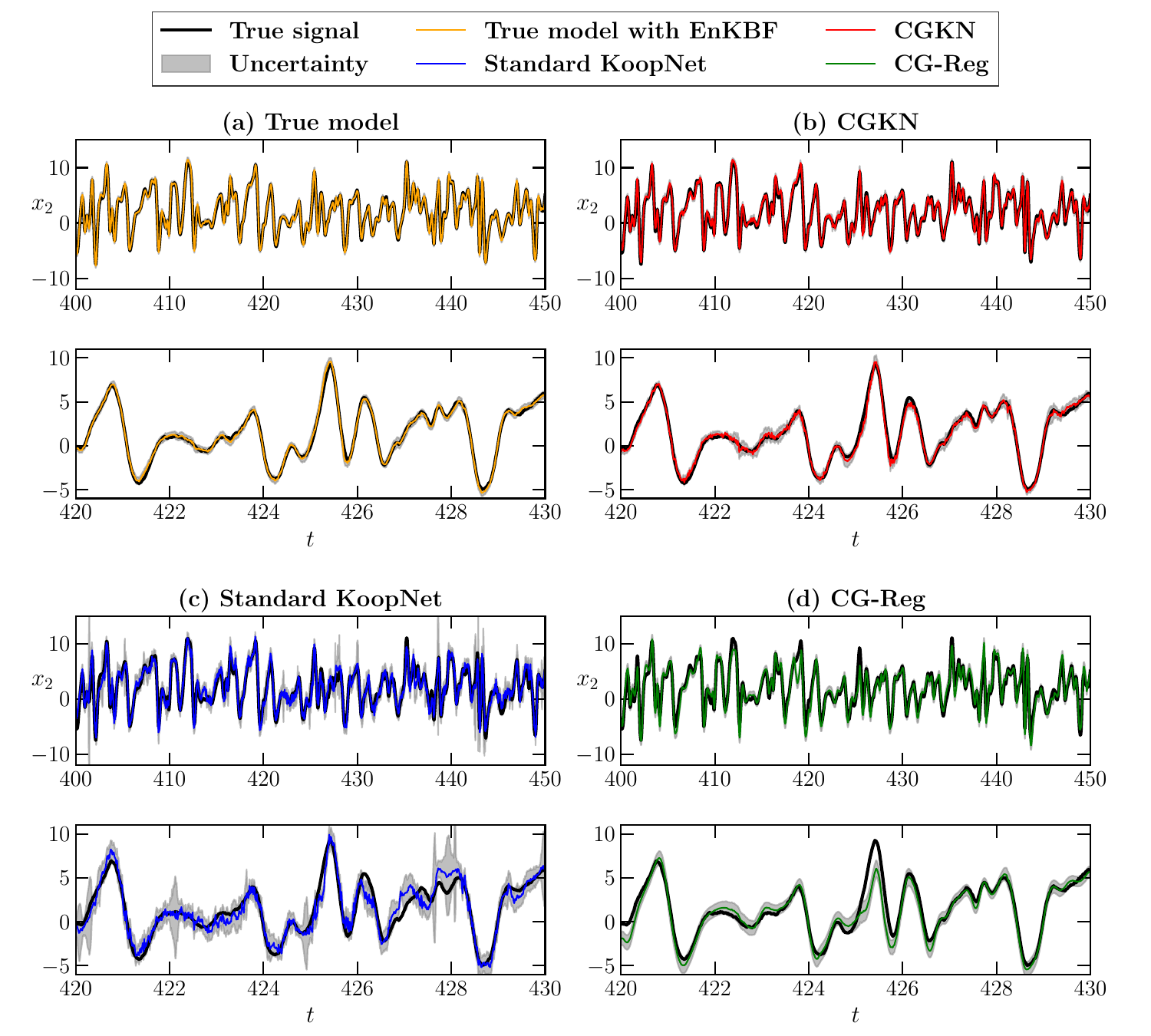}
    \caption{DA results of the unobserved state variable $x_2$ in the Lorenz 96 system. EnKBF is used for the true model, and analytic DA formulae are used for the other three models. The uncertainties are indicated by the grey-colored regions, which correspond to two estimated standard deviations from the posterior mean. The uncertainty area from the CGKN model and the standard KoopNet model are derived from residual analysis.}
    \label{fig:L96_DA}
\end{figure}

Figure~\ref{fig:L96_DA} presents the DA results of the unobserved state $x_2$. In each panel, the lower sub-figure provides a zoomed-in view of the upper one to clearly display the trajectory and uncertain area. In this case, the DA results from the true model with EnKBF yield a posterior mean that closely aligns with the true signal and a narrow uncertainty area, which is obtained from the estimated covariance of ensemble simulations. The CGKN model shows a good agreement with the true model and outperforms the standard KoopNet model and the CG-Reg model. For the CG-Reg model, the uncertainty area is derived from the covariance evolution formula with a similar form of Eq.~\eqref{eq:CGKN_CGF}. For both the CGKN model and the standard KoopNet model, the uncertainty area is based on the residual analysis method described in Section~\ref{SSec:CGKN_DA}.

\subsection{El Ni\~no-Southern Oscillation (ENSO): A multiscale non-Gaussian phenomenon}

The El Ni\~no-Southern Oscillation (ENSO) is the most significant source of interannual climate variability in the tropics, characterized by fluctuations in sea surface temperature (SST) in the equatorial Pacific \citep{ropelewski1987global, mcphaden2006enso, latif1998review, neelin1998enso}. Its warm (El Ni\~no) and cold (La Ni\~na) phases have profound effects on global weather patterns, ecosystems, and socioeconomic systems. Accurate forecasting of ENSO is crucial for improving climate predictions and mitigating the impact of extreme weather events.

A physics-based reference model is employed to generate synthetic ENSO data \citep{chen2023simple} for training and testing different methods here. This model consists of a set of stochastic partial differential equations (SPDEs) that describe the coupled dynamics of the atmosphere and ocean. It effectively captures many realistic features observed in the real world, including non-Gaussian statistics and variations in spatial patterns, peak intensity, and temporal evolution. Its capability to generate much longer time series than actual observations enables us to train and evaluate the skill of the CGKN on this complex natural phenomenon. More details and the governing equations of the ENSO SPDEs are available in \ref{Sec:ENSO_SPDEs}.

The physical variables of focus here are SST $\mathbf{T}$, thermocline depth $\mathbf{H}$, and wind amplitude $a_p$. The thermocline is a layer in a body of water where temperature decreases rapidly with depth, separating warmer surface waters from colder deep waters. SST and thermocline depth are considered interannual variables, evolving much more slowly than wind, which is classified as intraseasonal variability. SST is a commonly used indicator for measuring ENSO events, and it is strongly coupled with thermocline depth, together forming an irregular oscillator that reflects the interactions between the ocean and the sea surface. Despite its faster timescale, wind acts as an external forcing that drives the ocean and SST variables.

We conducted two simulations, each spanning 2000 years, using the ENSO SPDEs, and spatially averaged the state variables $\mathbf{T}$ (SST) and state $\mathbf{H}$ (thermocline depth) across three longitudinal ranges. The resulting discrete state variables are denoted as $T_W$, $T_C$, $T_E$ for the SST in the western, central, and eastern Pacific, respectively, along the equator, with the same notation for $H_W$, $H_C$ and $H_E$ for thermocline depth. It has been demonstrated that these coarse-grained state variables are sufficient to capture the large-scale features of ENSO dynamics \citep{chen2022multiscale, geng2020two}. Together with the state variable wind burst amplitude $a_p$, one 2000-year simulation of all $T$s and $H$s is used as training data and the other one is for testing, with both datasets having a time step of $\Delta t=1/360$ (about $1$ day).

\begin{figure}[H]
    \centering
    \includegraphics[width=\linewidth]{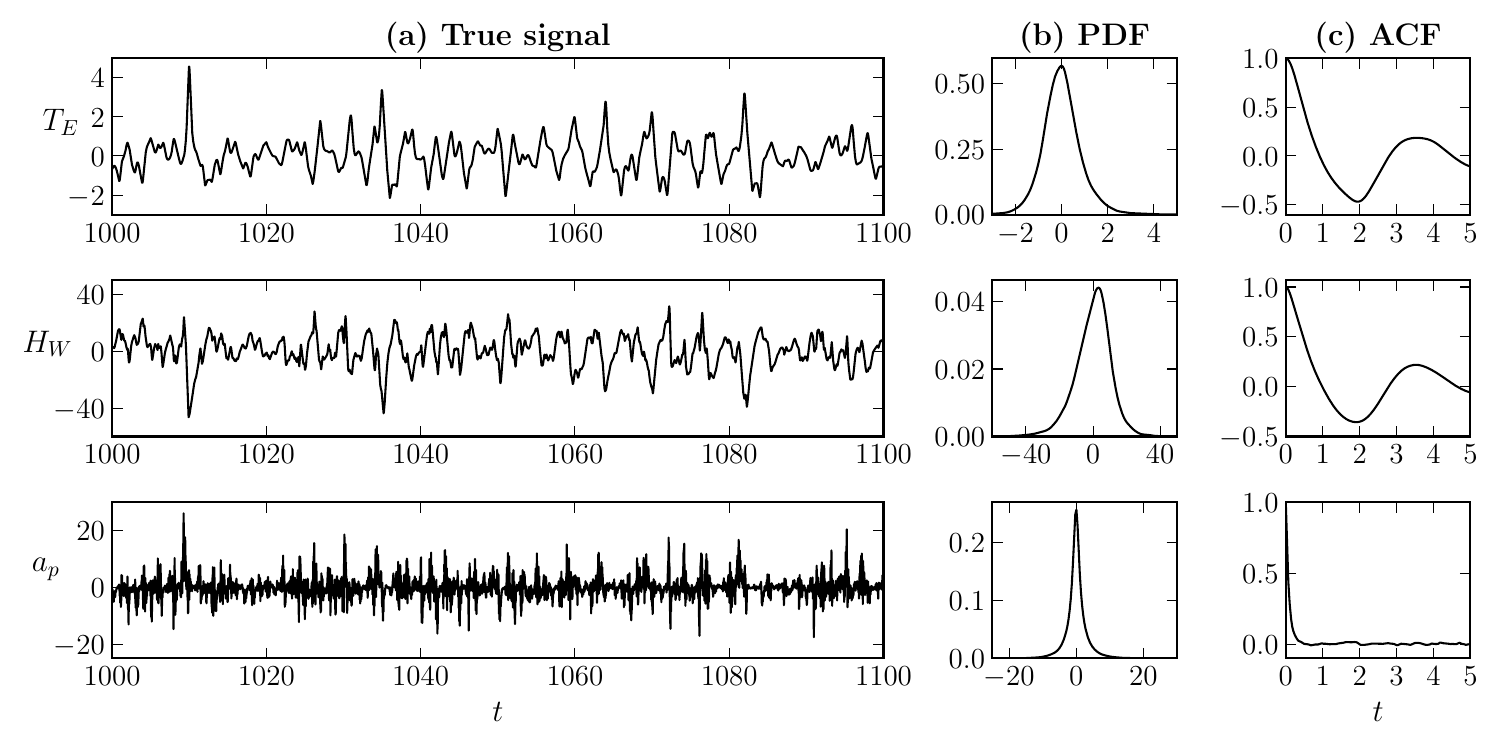}
    \caption{Time series and statistics property of states $T_E$, $H_W$, $a_p$ derived from the ENSO PDEs. Panel (a): part of time series of each state variable. Panel (b) and (c): probability density function (PDF) and auto-correlation function (ACF) of the corresponding state. It should be noted that the PDFs and ACFs are estimated from a much longer series than the one presented in panel (a).}
    \label{fig:ENSO_Property}
\end{figure}

Figure~\ref{fig:ENSO_Property} displays a portion of the training data for $T_E$, $H_W$, and $a_p$ in panel (a), along with their corresponding PDFs and ACFs in panels (b) and (c), evaluated using the entire training dataset. In this ENSO test case, the observed state variables are $T_W$, $T_C$, $T_E$, and $a_p$, while the unobserved ones are $H_W$, $H_C$, $H_E$, with $a_p$ serving as an external variable. The models, including CGKN, are trained on the data of all state variables and aim to forecast both $T$ and $H$ states, while performing DA without access to $H$ data. In practice, SST data is obtained from satellite instruments, while directly measuring thermocline depth is challenging. DA is involved in recovering the thermocline depth in practice. Therefore, estimating thermocline depth from SST in this setup simulates a realistic scenario.

With the encoder $\boldsymbol{\varphi}$, the decoder $\boldsymbol{\psi}$ and a neural network $\boldsymbol{\eta}$ that outputs $\mathbf{f}_1$, $\mathbf{g}_1$, $\mathbf{f}_2$, and $\mathbf{g}_2$, the CGKN model for this example is written as:
\begin{equation}
    \label{eq:ENSO_CGKN}
    \begin{aligned}
    \frac{\diff }{\diff t}[T_W, T_C, T_E, a_p]^\mathtt{T} &= {\mathbf{f}}_1(T_W, T_C, T_E, a_p) + {\mathbf{g}}_1(T_W, T_C, T_E, a_p)\mathbf{v} + \boldsymbol{\sigma}_1\dot{\mathbf{W}}_1, \\
    \frac{\diff \mathbf{v}}{\diff t} &= {\mathbf{f}}_2(T_W, T_C, T_E, a_p) + {\mathbf{g}}_2(T_W, T_C, T_E, a_p)\mathbf{v} + {\boldsymbol{\sigma}}_{\boldsymbol{v}}\dot{\mathbf{W}}_{\boldsymbol{v}},
    \end{aligned}
\end{equation}
where $\mathbf{v} = \boldsymbol{\varphi}(H_W, H_C, H_E)$ and $[H_W, H_C, H_E]^\mathtt{T} = \boldsymbol{\psi}(\mathbf{v})$. The latent state dimension of the autoencoder is set as 10 in this test case. With this choice, the sub-networks in the CGKN model are $\boldsymbol{\varphi}: \mathbb{R}^{3} \mapsto \mathbb{R}^{10}$, $\boldsymbol{\psi}: \mathbb{R}^{10} \mapsto \mathbb{R}^{3}$, $\boldsymbol{\eta}: \mathbb{R}^{4} \mapsto \mathbb{R}^{154}$ with number of parameters being 10554, 12970, 12963, respectively. In CGKN, the fully connected neural networks are used for encoder $\boldsymbol{\varphi}$, decoder $\boldsymbol{\psi}$, and sub-network $\boldsymbol{\eta}$, each consisting of six layers with ReLU activation functions. For the training settings, the state forecast time range $[t_0,t_{N_s}]$ is set as 1 year, and DA time range $[t_0,t_{N_l}]$ is 40 years with the warm-up period $t_{N_b}$ being 4 years. The weights in the target loss function are set as $\lambda_{\textrm{AE}} = 1/d_{\mathbf{u}_2}$, $\lambda_{\mathbf{u}} = 1/d_{\mathbf{u}}$, $\lambda_{\mathbf{v}}=0$ and $\lambda_{\textrm{DA}}=1/d_{\mathbf{u}_2}$. The gradient descent optimizer is selected as Adam with an initial learning rate of 1e-3 and a cosine annealing scheduler.

\begin{table}[H]
\caption{Test results of state forecast and data assimilation of all models for the example of ENSO. The errors are the normalized root mean squared error (NRMSE) between true values and approximated values.}
\label{tab:ENSO_Test_Errors}
\begin{adjustbox}{max width=1.\textwidth,center}
\begin{tabular}{|c|c|c|c|c|c|c|}
\hline
\diagbox{Models}{Errors} & Forecast Error & DA Error \\
\hline
True Model & 6.5406e-01 & --- \\
\hline
CGKN & 6.4762e-01 & 1.3518e-01  \\
\hline
Standard KoopNet & 6.8473e-01 & 2.3490e-01 \\
\hline
CG-Reg & 9.8751e-01 & NaN\\
\hline
DNN & 6.4845e-01 & ---  \\
\hline
\end{tabular}
\end{adjustbox}
\end{table}

The test results of all models for this case have been reported in Table~\ref{tab:ENSO_Test_Errors}. The ensemble number of the true model for ensemble mean state forecast is set to 100. The DA error from EnKBF has been omitted due to the computational resource required by the ENSO PDEs and it is not the focus of this work. For the performance of state forecast, except for the CG-Reg model, the other models achieve similar results with the standard KoopNet model ranked the lowest. For the DA performance, the CG-Reg model becomes unstable and diverges over the horizon of 2000 years from test data. The DA error of the CGKN model is the lowest, outperforming the performance of the standard KoopNet. In this test case, the CGKN model demonstrates optimal performance in both state forecast and DA compared to all other models.

\begin{figure}[H]
    \centering
    \includegraphics[width=0.9\linewidth]{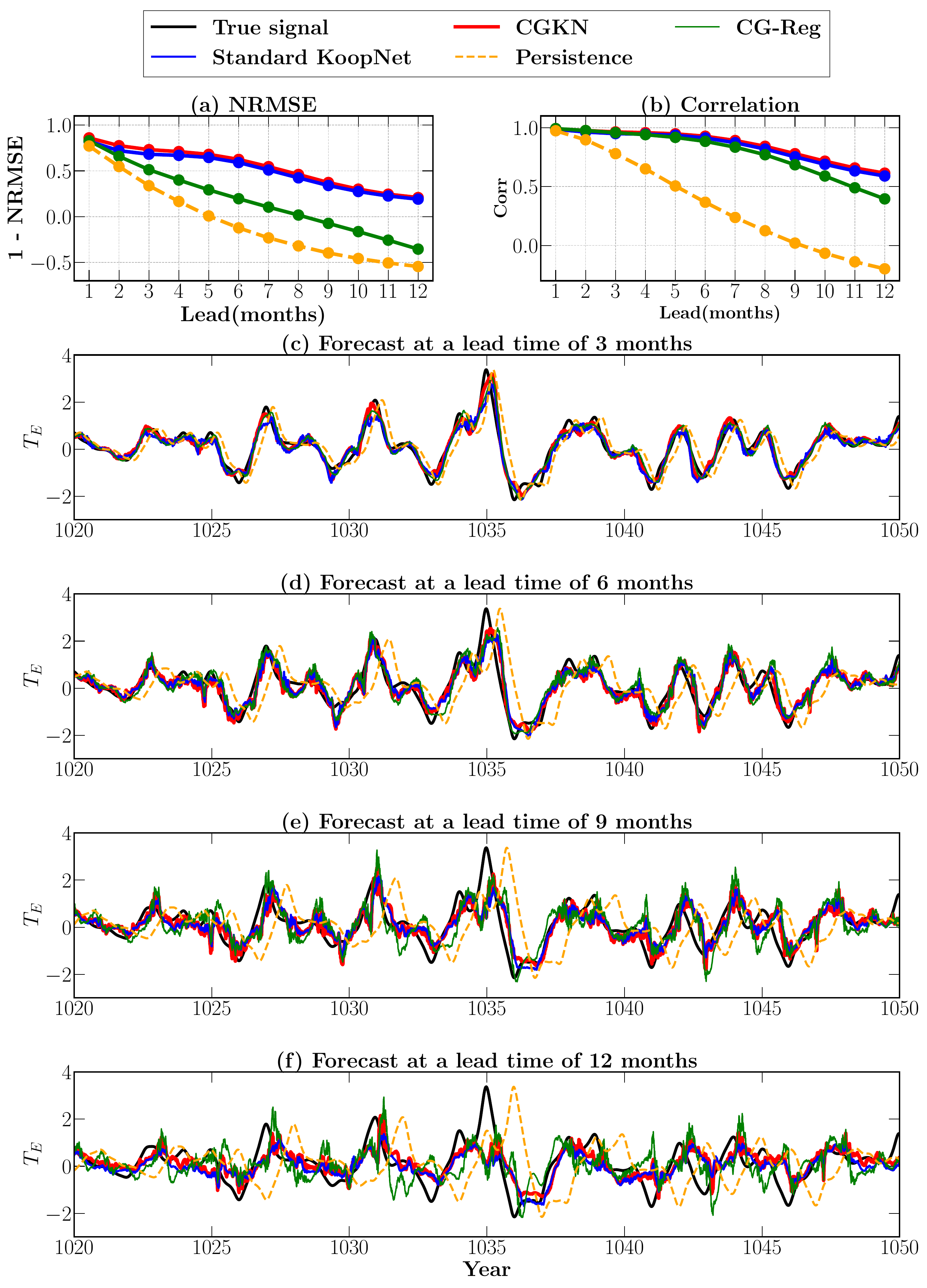}
    \caption{Comparison of the lead-time forecast for state $T_E$ from different models. Panel (a) and (b): skill scores including normalized root mean squared error (NRMSE) and correlation of the CGKN model, standard KoopNet model, CG-Reg model, and Persistence. Panel (c) - (f): comparison of the true signal and forecast results from different models at lead times of 3, 6, 9, and 12 months, respectively, over the years 1020 to 1050.}
    \label{fig:ENSO_LeadForecast}
\end{figure}

Figure~\ref{fig:ENSO_LeadForecast} shows the lead time prediction of different models in terms of $T_E$, which is one of the most widely used ENSO indicator. The panels (a) and (b) display the NRMSE and the pattern correlation between the true signal of state $T_E$ and its predictive states at different lead times. The correlation is a standard measurement in climate science and it is therefore included here. The NRMSE is calculated based on the formula of Eq.~\eqref{eq:NRMSE} and the correlation is from the formula:
\begin{equation}
    \label{eq:corr}
    \textrm{Corr} = \frac{\sum_{n=1}^{N} \big(x^{\star}(t_n) - \bar{x}^{\star} \big) \big(x(t_n) - \bar{x}\big)}{\sqrt{\sum_{n=1}^{N} \big(x^{\star}(t_n) - \bar{x}^{\star}\big)^2}\sqrt{\sum_{n=1}^{N} \big(x(t_n) - \bar{x}\big)^2}}
\end{equation}
with $x^{\star} \in \mathbb{R}^N$ is the true values of state time series, $x$ is the approximated values, $x(t_n)$ is the state at time $t_n$. The $\bar{x}^{\star}$ and $\bar{x}$ are the time average of true values and approximated values. In this example, the persistence model, a simple and commonly used prediction method for ENSO, is considered. It assumes that ENSO conditions remain unchanged over time and use the current values as future predictions. The persistence serves as a baseline model for evaluating other computational models. The panels showing NRMSE and correlation indicate that both the CGKN model and the standard KoopNet model produce stable and accurate state forecast results across various lead times, demonstrating the effectiveness of these deep learning models. In contrast, the CG-Reg model exhibits a rapid decline in forecast performance due to its omission of certain key dynamics in the parametric forms, while the persistence model performs the worst among all the models. Panels (c)-(d) show the time series of the true state and predictive states at a lead time of 3 months, 6 months, 9 months, and 12 months, respectively. For the lead time prediction at 3 months in panel (c), all the models including persistence can almost follow the true signal. However, with the lead time becoming longer, the deviation and shift pattern of predictions from all models have become more pronounced. For the 12-month lead time prediction in panel (f), although the forecast skill of the CGKN model declines, it still performs the best overall. The model maintains a correlation above 0.5 and significantly outperforms the persistence model.

\begin{figure}[H]
    \centering
    \includegraphics[width=0.9\linewidth]{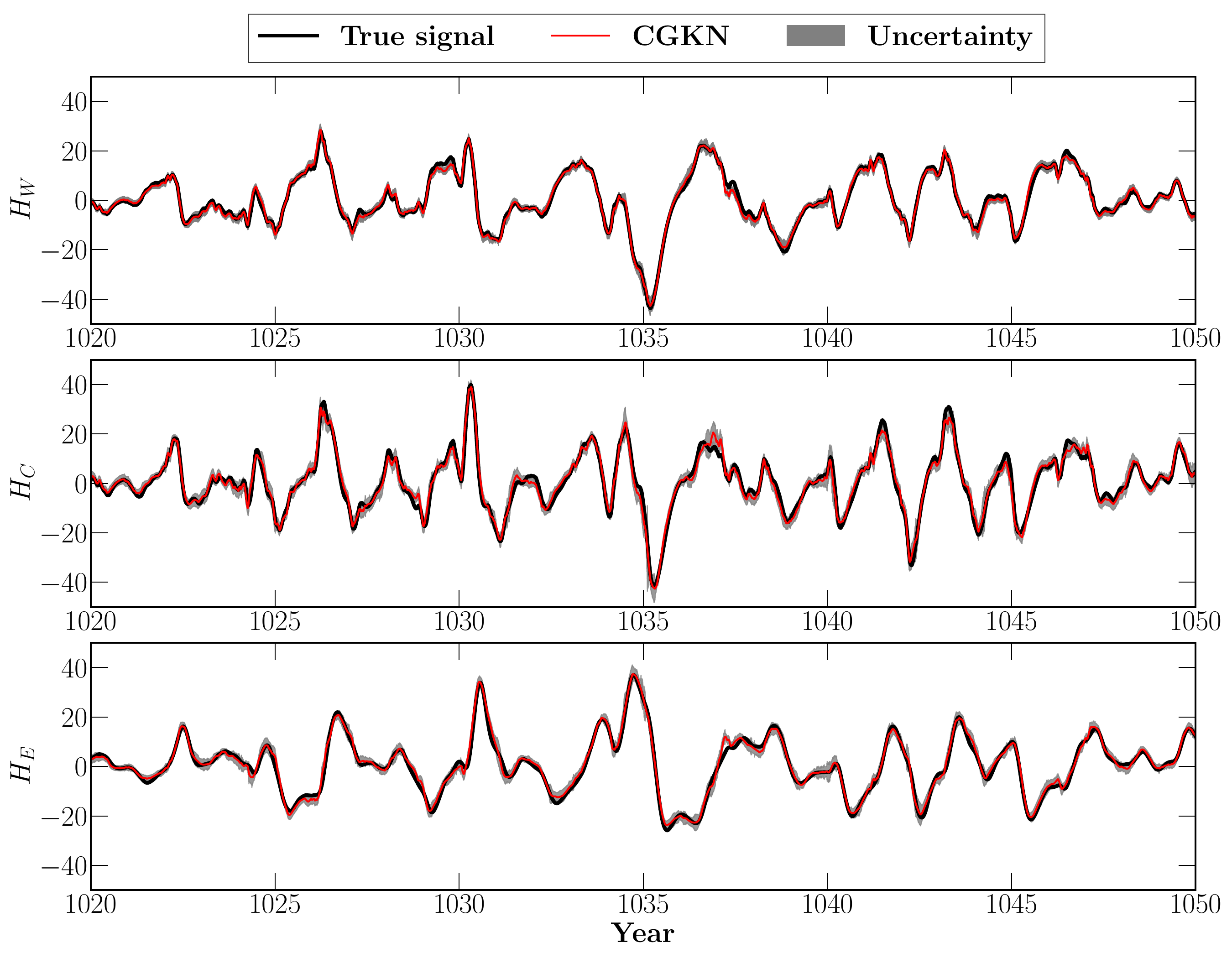}
    \caption{DA results of the unobserved state variables $H_W$, $H_C$, and $H_E$ derived from physics-based PDEs for ENSO. The DA posterior mean is obtained from applying the analytic DA formulae to the CGKN model and the uncertainty is estimated based on the method of residual analysis. The uncertainties are indicated by the grey-colored regions, which correspond to two estimated standard deviations from the posterior mean.}
    \label{fig:ENSO_DA}
\end{figure}

Figure~\ref{fig:ENSO_DA} shows the DA results from the CGKN model for estimating the unobserved state $H_W$, $H_C$ and $H_E$ given trajectories of observed state $T_W$, $T_C$, $T_E$ and $a_p$. 
The governing equations of the true model for ENSO are stochastic partial differential equations (SPDEs), detailed in Appendix B. The SPDEs are computationally intensive and time-consuming for ensemble methods to perform data assimilation effectively. For the CG-Reg model, the DA result is very unstable and diverges quickly, leading to the finite-time blow-up of the posterior mean and covariance. In addition, the standard KoopNet shows a similar DA result to the CGKN. Therefore, Figure~3.8 primarily focuses on the DA results of CGKN compared to the true signal.  
The DA posterior mean from the CGKN model is obtained by applying the analytic formulae in Eq.~\eqref{eq:CGKN_CGF}, while the uncertainty areas are determined using the method of residual analysis described in Section~\ref{SSec:CGKN_DA}. The similarity between the posterior mean from the CGKN model and the true signal confirms its capability in DA. Although the posterior mean from the CGKN model shows does not perfectly match the true signal occasionally, the uncertainty area can still cover the true data for most of those times.

\begin{figure}[H]
    \centering
    \includegraphics[width=1\linewidth]{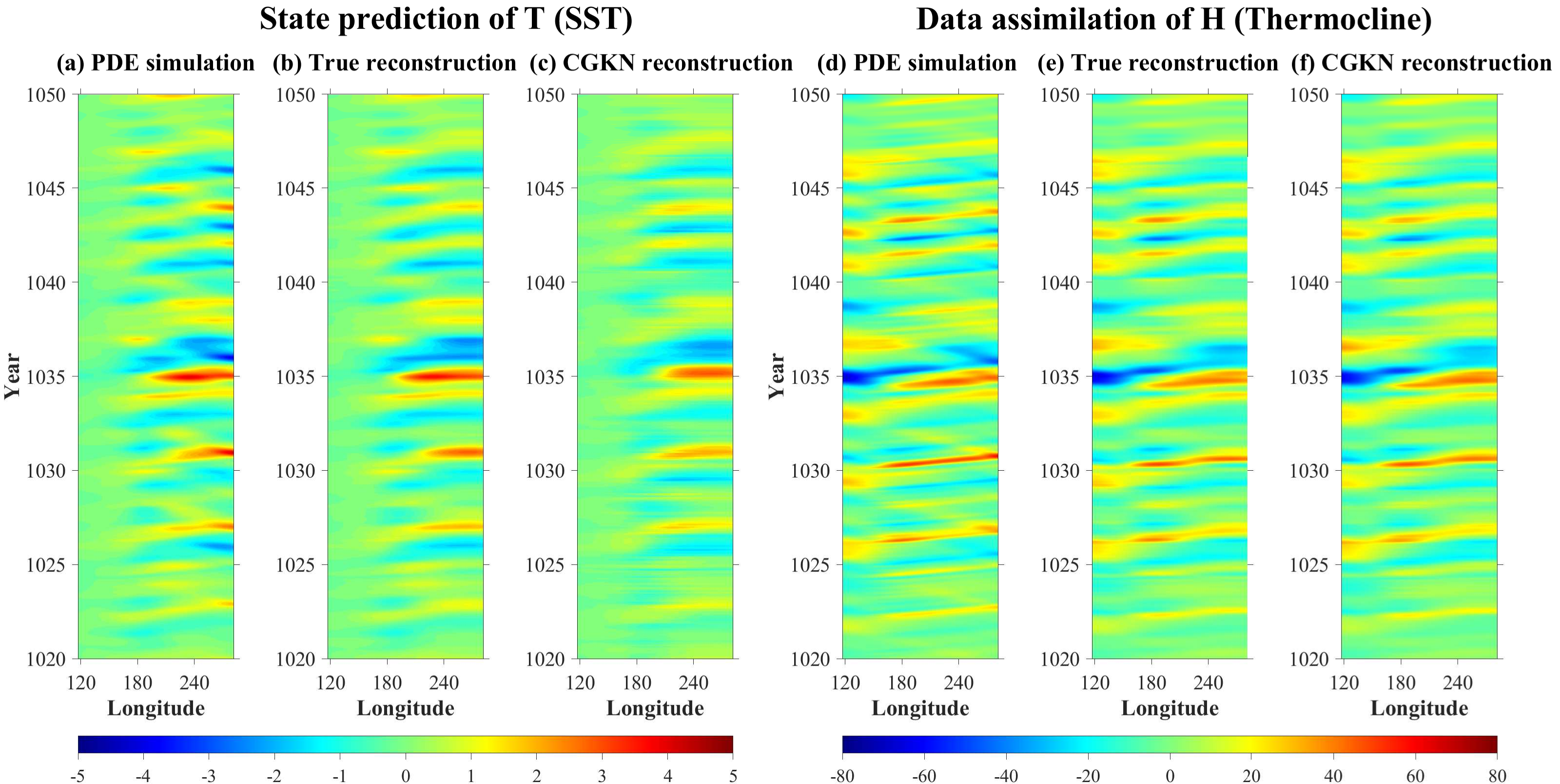}
    \caption{Hovmoller diagrams for the simulation from true model and reconstructions by true and approximated states. Panel (a): simulation of $\mathbf{T}$ (SST) from the true model. Panel (b): reconstruction of $\mathbf{T}$ from the true discrete states $T_W$, $T_C$ and $T_E$ which are spatial average of $\mathbf{T}$. Panel (c): reconstruction of $\mathbf{T}$ from predictive states $T_W$, $T_C$ and $T_E$ of the CGKN model at a 6-month lead time. Panel (d): simulation of $\mathbf{H}$ (thermocline) from the true model. Panel (e): reconstruction of $\mathbf{H}$ from the true discrete states $H_W$, $H_C$ and $H_E$ which are spatial average of $\mathbf{H}$. Panel (f): reconstruction of $\mathbf{H}$ from estimated $H_W$, $H_C$, and $H_E$ via DA which apply analytical formulae on CGKN. }
    \label{fig:ENSO_Hov}
\end{figure}

In addition to comparing the trajectories of the true states and approximated ones from forecast and DA, the spatiotemporal simulation of $\mathbf{T}$ (SST) and $\mathbf{H}$ (thermocline) of the true fields and the approximated ones are also presented in Fig.~\ref{fig:ENSO_Hov}. The approximated fields are obtained from the reconstruction of discrete states $T$s and $H$s via a bi-variate regression method:
\begin{equation}
\label{Eq:reg}
\begin{aligned}
\tilde{\mathbf{T}}(x, t) & =r_W(x) T_W(t)+r_C(x) T_C(t)+r_E(x) T_E(t),\\
\tilde{\mathbf{H}}(x, t) & =l_W(x) H_W(t)+l_C(x) H_C(t)+l_E(x) H_E(t),
\end{aligned}
\end{equation}
where $x$ is the longitude and $t$ is the temporal variable. The regression coefficients $r_W(x)$, $r_C(x)$, $r_E(x)$, $l_W(x)$, $l_C(x)$, and $l_E(x)$ are determined using the true signals at each grid point of longitude $x$. The discrete states from either the spatial average of true simulation or inference of the CGKN model are plugged into the regression formula in Eq.~\eqref{Eq:reg} to obtain the reconstructed SST field $\mathbf{T}$ and reconstructed thermocline field $\mathbf{H}$. Figure~\ref{fig:ENSO_Hov} shows the spatial-temporal evolution of true simulations, true reconstruction from spatial-averaged state $T$s and $H$s, and reconstruction from approximated ones via the CGKN model. In the third column of left panel, the predictive spatial-temporal simulation of $\mathbf{T}$ is reconstructed from 6-month lead time forecast of $T_W$, $T_C$ and $T_E$ via CGKN.  The corresponding time series of $T_E$ is shown in Fig.~\ref{fig:ENSO_LeadForecast}(d). The comparison to the true PDE simulation confirms the capability of state forecast of the CGKN model since the overall pattern of the evolution of SST can be reproduced. The DA results are shown in the right panel. The spatiotemporal simulation of $\mathbf{H}$ (thermocline) in the third column is reconstructed from the estimated states $T_W$, $T_C$, and $T_E$ obtained by applying the analytic DA formulae in Eq.~\eqref{eq:CGKN_CGF} to the CGKN model. With these accurate DA posterior means of states $H$s as shown in Fig.~\ref{fig:ENSO_DA}, the successful reconstruction of the spatially continuous $\mathbf{H}$ field is as expected.

\section{Discussion and Conclusion}
\label{Sec:Conclusion}
In this work, we propose a conditional Gaussian Koopman network --- CGKN --- as a unique deep learning framework for modeling complex dynamical systems that facilitate both accurate forecasts and efficient DA. The proposed framework is inspired by the Koopman theory. It builds upon the conditional Gaussian nonlinear system, transforming any nonlinear system into a modeled system that embeds a conditional linear structure to facilitate efficient analytic formulae of DA. The proper nonlinear transformation is jointly learned with the unknown terms in the transformed conditional Gaussian nonlinear system, based on a total loss function that accounts for the overall performance of the nonlinear transformation, the state prediction, and the DA. Among all the models that support efficient analytic formulae of DA, the numerical examples demonstrate that the CGKN model generally outperforms the standard KoopNet model and the CG-Reg model in both state prediction and DA.

The proposed framework integrates the strengths of deep learning methods, Koopman theory, and conditional Gaussian nonlinear systems for modeling complex dynamical systems and efficient DA. In addition, it serves as an effort towards a systematic approach to impose useful structures into deep learning models and to facilitate their usage in other outer-loop applications of scientific machine learning, e.g., control of partially observed systems, and inverse problems. Some extensions and future directions of the proposed framework include more rigorous approaches to efficiently quantify the uncertainties associated with the DA results of the CGKN model and multi-objective optimization methods to enhance the overall performance of the nonlinear transformation, the state predictions, and the DA.

\section*{Acknowledgments}

The research of C.C. and J.W. was funded by the University of Wisconsin-Madison, Office of the Vice Chancellor for Research and Graduate Education, with funding from the Wisconsin Alumni Research Foundation. The research of N.C. is funded by the Office of Naval Research N00014-24-1-2244. Y.Z. is partially supported as a research assistant under this grant.

\section*{Data Availability}

The data that support the findings of this study are available from the corresponding author upon reasonable request. The codes and examples that support the findings of this study are available in the link: \url{https://github.com/ChuanqiChenCC/CGKN}.

\bibliographystyle{unsrt}
\bibliography{references}

\clearpage

\appendix

\section{Additional Results of the Projected Stochastic Burgers–Sivashinsky Equation}
\label{Sec:PSBSE_Extra}

Table~\ref{tab:Test_Errors2} summarizes some additional results of the projected stochastic Burgers-Sivashinsky equation, for comparing the performance of CGKN and other methods with different choices of observed and unobserved state variables. In addition to the setup of state variable $x$ being observed as studied in Section~\ref{Sec:NumericalExperiments}, we also investigate two different setups with state variables $y$ and $z$ being observed, respectively. The results in Table~\ref{tab:Test_Errors2} confirm that the performance of CGKN is not sensitive to the specific choice of observed and unobserved states presented in Section~\ref{Sec:NumericalExperiments}.

\begin{table}[H]
\caption{Test results of state forecast and data assimilation under different setups of observed and unobserved states for the example of projected stochastic Burgers–Sivashinsky equation. The errors are the normalized root mean squared error (NRMSE) between true values and approximated values.}
\label{tab:Test_Errors2}
\begin{adjustbox}{max width=1.\textwidth,center}
\begin{tabular}{|c|c|c|c|c|c|c|}
\hline
\multirow{2}{*}{\diagbox{Models}{Errors}} & \multicolumn{2}{c|}{Obs:$x$ ~~~ Unobs:$y, z$} & \multicolumn{2}{c|}{Obs:$y$ ~~~ Unobs:$x, z$} & \multicolumn{2}{c|}{Obs:$z$ ~~~ Unobs:$x, y$} \\
\cline{2-7}
& Forecast Error & DA Error & Forecast Error & DA Error & Forecast Error & DA Error \\
\hline
True Model & 2.4566e-01 & 6.9466e-01 & 2.4566e-01 & 9.7836e-01 & 2.4566e-01 & 1.4051e+00 \\
\hline
CGKN & 2.8389e-01 & 7.2776e-01  & 2.9901e-01  & 1.0992e+00  & 2.9863e-01 & 8.4302e-01 \\
\hline
Standard KoopNet & 3.7600e-01 & 7.5069e-01 & 4.1460e-01 & 1.3753e+00 & 3.3586e-01 & 8.8313e-01 \\
\hline
CG-Reg & 2.7766e-01 & 1.2884e+00   & 3.3051e-01 & 1.3919e+00 & 4.4084e-01 & 6.5044e+00 \\
\hline
DNN & 2.4728e-01 & --- & 2.4728e-01 &  --- & 2.4728e-01 & --- \\
\hline
\end{tabular}
\end{adjustbox}
\end{table}

\section{A Quick Summary of the ENSO SPDE Test Model}
\label{Sec:ENSO_SPDEs}

The physics-based reference ENSO model \citep{chen2023simple} discussed in this paper is built based on the dynamics of atmosphere-ocean interactions in the tropical Pacific. The model consists of two main components: a deterministic dynamical core that captures the essential physics of ENSO and stochastic parameterizations that trigger the extreme events showing the atmospheric and oceanic processes occurring at different timescales.

The dynamical core describes the interactions between the atmosphere, the ocean, and the SST through three key mechanisms. First, the atmosphere drives ocean circulation through wind stress, where the wind busts generate currents and affect the thermocline depth. Second, the ocean affects the atmosphere through latent heat, which is proportional to SST. Third, the ocean modulates SST through changes in the thermocline depth and zonal currents.

The atmospheric component is based on a non-dissipative Matsuno-Gill type atmosphere model ~\citep{gill1980some,matsuno1966quasi}. This model is a simplified representation of atmospheric dynamics in the tropics, designed to capture large-scale motions. A key feature in the model of Eq.~\eqref{ENSO:atm} is the latent heating $E_q$, which is proportional to the sea surface temperature and drives the atmospheric circulation. For the ocean component, the model employs a simple shallow-water system ~\citep{vallis2016geophysical}. The model focuses on the dynamics of an active upper layer. The upper layer dynamics are driven by surface wind stress from the atmosphere. This wind stress causes currents and waves, representing momentum transfer from the atmosphere to the ocean. These assumptions result in the reduced shallow-water equations shown in Eq.~\eqref{ENSO:Ocean}. The SST equation is derived from an SST budget equation ~\citep{jin1997equatorial}. These include vertical temperature changes due to thermocline variations, horizontal heat transport by ocean currents, and thermal damping through air-sea heat exchange. These essential mechanisms are captured in Eq.~\eqref{ENSO:SST}.

The dynamical core is a coupled atmosphere-ocean-SST system:

Atmosphere:

\begin{equation}\label{ENSO:atm}
\begin{aligned}
& -y v-\partial_x \theta=0 \\
& y u-\partial_y \theta=0 \\
& -\left(\partial_x u+\partial_y v\right)=E_q /(1-\bar{Q})
\end{aligned}
\end{equation}

Ocean:

\begin{equation}\label{ENSO:Ocean}
\begin{aligned}
& \partial_t U-c_1 Y V+c_1 \partial_x H=c_1 \tau_x \\
& Y U+\partial_Y H=0 \\
& \partial_t H+c_1\left(\partial_x U+\partial_Y V\right)=0
\end{aligned}
\end{equation}

SST:

\begin{equation}\label{ENSO:SST}
\partial_t T=-c_1 \zeta E_q+c_1 \eta_1 H+c_1 I \eta_2 U .
\end{equation}

In the atmosphere component \eqref{ENSO:atm}, $u$, $v$, and $\theta$ represent the zonal wind speeds, meridional wind speeds, and the potential temperature, respectively. In the ocean components \eqref{ENSO:Ocean}, $U$, $V$ are zonal and meridional ocean currents, $H$ is the thermocline depth. Besides, $T$ represents the SST. These variables are all anomalies. The $t$ represents the interannual time coordinate, $x$ is the zonal coordinate, and $y$ and $Y$ are the meridional components for the atmosphere and ocean components. The equations include various parameters such as latent heat $E_q = \alpha_q T$, wind stress $\tau_x$, latent heating exchange coefficient $\zeta$, and the strengths of the thermocline $\eta_1$ and zonal advective feedback $\eta_2$. The relative magnitudes of $\eta_1$ and $\eta_2$ differ between the Eastern Pacific (EP) and Central Pacific (CP) due to variations in thermocline depth and zonal SST gradients. The model assumes periodic boundary conditions for the atmosphere and reflection boundary conditions for the Pacific Ocean. Since ENSO is primarily a phenomenon near the equator, we further project the model to the leading basis in the meridional direction.

While this deterministic dynamical core captures the basic ENSO physics, additional processes at different timescales play crucial roles in generating realistic features of ENSO diversity. These additional effects can be effectively characterized by stochastic parameterizations, which describe the intraseasonal wind variations and the decadal changes in the background conditions. 

The wind stress $\tau_x$ consists of two components, the atmospheric circulation $u$ and the stochastic wind bursts $u_p$, which is localized in the western Pacific (WP) and has the following structure:
\begin{equation}
u_p(x, y, t) = a_p(t) s_p(x) \phi_0(y)
\end{equation}
where $\phi_0(y)$ is the leading meridional basis, $s_p(x)$ is a fixed spatial structure, and $a_p(t)$ is the wind burst amplitude governed by a stochastic process:
\begin{equation}
\frac{\mathrm{d} a_p}{\mathrm{~d} t} = -d_p a_p + \sigma_p(T_C) \dot{W}_p
\end{equation}
Here, $d_p$ is the damping term, $\dot{W}_p$ is a white noise source, and $\sigma_p(T_C)$ is the state-dependent noise strength, which is a function of the SST averaged over the CP. In the absence of seasonal cycle and decadal influence, the noise strength is parameterized as $\sigma_p(T_C) = 1.6(\tanh(T_C) + 1)$.
This state-dependent noise coefficient implies that wind burst activity is more active during El Ni\~no events due to the eastward extension of the warm pool. The intraseasonal wind bursts not only affect the interannual variability but are also modulated by it.

Finally, the decadal variability is driven by a simple stochastic process:
\begin{equation}
\frac{\mathrm{d} I}{\mathrm{~d} t} = -\lambda(I - m) + \sigma_I(I) \dot{W}_I,
\end{equation}
where $\lambda$ is the damping term set to 5 years$^{-1}$, representing the decadal time scale. $\sigma_I(I)$ is the state-dependent noise strength, and $\dot{W}_I$ is the white noise source. The state-dependent noise coefficient allows the distribution of $I$ to be non-Gaussian, with $I$ representing the strength of the easterly trade wind in the lower level of the Walker circulation on the decadal time scale. A uniform distribution between $[0, 1]$ is adopted for $I$, with larger values corresponding to stronger easterly trade winds.

\end{document}